\documentclass[11pt]{article}
\usepackage[numbers]{natbib}
\usepackage{EMNLP2023}

\usepackage{times}
\usepackage{latexsym}
\usepackage{amsmath,amsfonts,amsthm,amssymb}
\usepackage{bm}
\usepackage{algorithmic}
\usepackage{algorithm}
\usepackage{array}
\usepackage[caption=false,font=normalsize,labelfont=sf,textfont=sf]{subfig}
\usepackage{textcomp}
\usepackage{stfloats}
\usepackage{url}
\usepackage{verbatim}
\usepackage{graphicx}
\usepackage{tabularx}
\usepackage{adjustbox}
\usepackage[T1]{fontenc}

\usepackage[utf8]{inputenc}

\usepackage{microtype}

\usepackage{inconsolata}

\title{Transformers and Language Models in Form Understanding: A Comprehensive Review of Scanned Document Analysis}

 \author{Abdelrahman Abdallah \and Daniel Eberharter \and Zoe Pfister \and  Adam Jatowt \\
         Department of Computer Science, \\ University of Innsbruck \\ Innsbruck}
\begin{document}
\maketitle
\begin{abstract}
This paper presents a comprehensive survey of research works on the topic of form understanding in the context of scanned documents. We delve into recent advancements and breakthroughs in the field, highlighting the significance of language models and transformers in solving this challenging task. Our research methodology involves an in-depth analysis of popular documents and forms of understanding of trends over the last decade, enabling us to offer valuable insights into the evolution of this domain. Focusing on cutting-edge models, we showcase how transformers have propelled the field forward, revolutionizing form-understanding techniques. Our exploration includes an extensive examination of state-of-the-art language models designed to effectively tackle the complexities of noisy scanned documents. Furthermore, we present an overview of the latest and most relevant datasets, which serve as essential benchmarks for evaluating the performance of selected models. By comparing and contrasting the capabilities of these models, we aim to provide researchers and practitioners with useful guidance in choosing the most suitable solutions for their specific form understanding tasks.
\end{abstract}

\section{Introduction}
Digitization has become ubiquitous in various aspects of our lives, where physical objects like music discs and analog documents are being replaced with digital, machine-readable counterparts. This transformation enables downstream applications and historical preservation through efficient storage in databases~\cite{davis2019deep,abdallah2022tncr,zhong2019publaynet}.

Among the crucial tasks in digitization, extracting form information from scanned documents stands out as a critical challenge. Digitizing form documents is non-trivial due to their inherent complexity, including intricate structures, non-textual elements like graphics, and combinations of hand- and machine-written content~\cite{Lewis2006BuildingAT,nurseitov2021handwritten,nurseitov2021classification}. Moreover, scanned documents may suffer from noise, resulting in unclear or distorted images, further complicating the form extraction process. This survey paper focuses on the methodologies employed for extracting form information from noisy scanned documents.

Document understanding~\cite{nurseitov2021classification,kim2022ocrfree,toiganbayeva2022kohtd,kasem2022deep} is the field of analyzing the content and structure of documents with various formats and modalities, such as text, images, tables, and graphs. Language models and transformers~\cite{irie2019language} are advanced neural network architectures that have demonstrated remarkable capabilities in various natural language processing tasks, as well as computer vision and audio processing tasks. The transformative potential of these models in document understanding has been widely recognised~\cite{wang2019language}.

\begin{figure*}
\centering
\includegraphics[width=.95\linewidth]{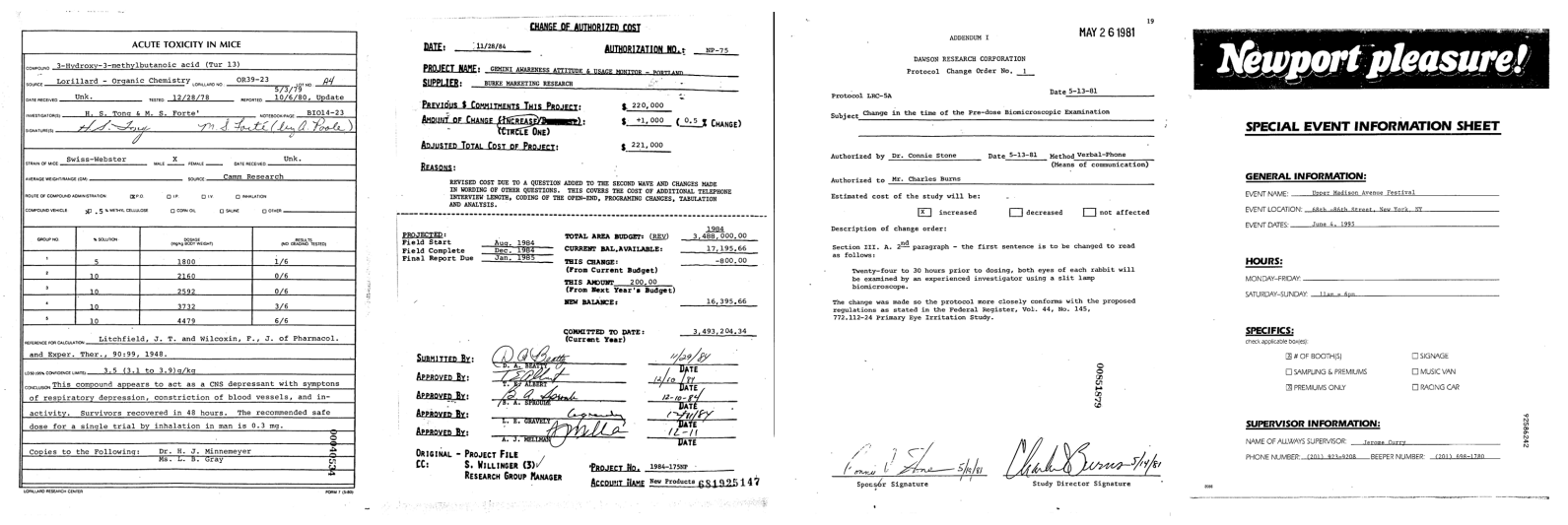}
\caption{Examples of scanned documents with different layouts~\cite{Xu2020LayoutLMPO}.}
\label{fig:document_examples}
\end{figure*}

Form understanding is essentially a sequence-labeling task akin to named entity recognition (NER), often referred to as \emph{key information extraction} (KIE)~\cite{abdallah2023amurd}. The task poses unique challenges due to the diverse shapes and formats in which these documents appear (see Fig~\ref{fig:document_examples} for examples). Unlike traditional NER tasks that deal with 1D text information, form understanding involves multiple modalities. In addition to textual information, the position and layout of text segments play a vital role in semantic interpretation. An effective language model for form understanding must comprehend document entities within the context of multiple modalities and adapt to various types of document formatting.

One of the challenges in form understanding is the integration of visual information from scanned documents. Vision-language models, such as ViT (Vision Transformer) \cite{9716741,wu2021cvt} and CLIP (Contrastive Language–Image Pretraining)\cite{li2021supervision} have demonstrated the potential to bridge the gap between text and images. These models enable the understanding of visual elements within the context of textual content, offering new ways for improved form understanding. Transformer-based models \cite{devlin2018bert,radford2019language} belong to a type of deep learning approach that is very effective in a wide range of natural language processing~\cite{abdallah2023generator,abdallah2023exploring} and computer vision~\cite{han2022survey,arnab2021vivit} tasks. In the domain of form understanding, Transformer models have been particularly transformative. They excel in interpreting the complex, often unstructured data found in forms and documents, thanks to their ability to capture contextual relationships in text. For instance, BERT's bidirectional training mechanism enables it to understand the context of a word based on its surrounding text, which is crucial in deciphering the semantics of information in forms.  
The key feature of these models is the use of self-attention mechanisms, which enables them to effectively weigh the importance of different parts of the input when making predictions. This allows for understanding the context of a given question and providing a relevant answer.

A notable example of the fusion of language models and transformers in document understanding is the LayoutLM model~\cite{Xu2020LayoutLMPO}. This model extends BERT by incorporating layout information, enabling it to consider the spatial arrangement of tokens in addition to their textual context. LayoutLM demonstrated superior performance in tasks like document image classification and form understanding, where both textual and spatial information are crucial.

It is a challenging and important problem in the field of document analysis and understanding. While there are several surveys that cover general aspects of document understanding and layout analysis~\cite{binmakhashen2019document,subramani2020survey}, as well as specific topics such as visual document analysis~\cite{igorevna2022document}, Camera-based analysis of text and documents~\cite{liang2005camera}, and deep learning for document image understanding~\cite{shigarov2023table}, there is currently no comprehensive survey focusing specifically on the task of \textit{form understanding}. Our survey bridges this gap and provides an extensive overview of recent technologies dedicated to the form understanding task. 

The main contributions of this paper can be summarized as follows:

\begin{enumerate}
    \item Comprehensive survey of recent advancements in form understanding for scanned documents, highlighting the role of language models and transformers.
    
    \item Rigorous methodology for literature collection, encompassing historical evolution of form understanding methodologies and specialized techniques.
    
    \item In-depth analysis of cutting-edge transformer models, focusing on their effectiveness in handling noisy scanned documents.
    
    \item Overview of essential datasets serving as benchmarks for evaluating model performance. Identification of research gap and contribution through a dedicated form understanding survey.
\end{enumerate}
The structure of our paper is as follows: In Section~\ref{sec:methodology}, we present our research methodology, detailing how we collected and curated papers for this survey. Section~\ref{sec:previous-work} provides an in-depth exploration of previously employed methods in the domain of form and document understanding. We delve into different approaches to address form understanding in noisy scanned documents in Section~\ref{sec:approach-multi-modal}. In Section~\ref{sec:datasets}, we describe commonly used datasets, which serve as the foundation for comparing various approaches. The comparative analysis is presented in Section~\ref{sec:comparison}. Finally, we offer an outlook on potential future directions and conclude in Section~\ref{sec:conclusion}.

\section{Literature Collection Methodology}
\label{sec:methodology}

To ensure the comprehensiveness of our survey, we employed a rigorous methodology for collecting relevant literature in the field of form understanding in noisy scanned documents. Our goal was to cover recent advancements and capture the latest trends in this rapidly evolving domain.

We conducted searches in prominent scientific databases, namely Scopus, on Aug 6th, 2023. The search query was designed to be comprehensive, combining relevant terms to target the topic effectively: "document understanding" AND "form" OR "information extraction" AND "Invoices"). By incorporating \texttt{"document understanding" AND "form"} in the search string, we ensured a strong connection between the topics of form understanding and document understanding.

The search results revealed substantial publication numbers across the databases. Springer Link returned the highest number of publications (162), with 50 of them published in 2021 alone. ACM yielded 75 relevant publications, with a noticeable increase in publications after 2018. IEEE returned 8 relevant publications.

In addition to the initial search results, we included the LayoutLMv2 approach~\cite{Xu2021LayoutLMv2MP} as it represented one of the state-of-the-art models during our survey's timeline.

To ensure the thoroughness of our literature collection, we employed the backward-snowballing method on the initially collected papers. This approach helped us identify additional relevant papers that may not have appeared in the initial search results. Through this process, we selected 
the publications that showcased the latest advances in form understanding and introduced datasets dedicated to form understanding or document understanding as a whole.

\begin{figure}[h]
\centering
\includegraphics[width=\linewidth]{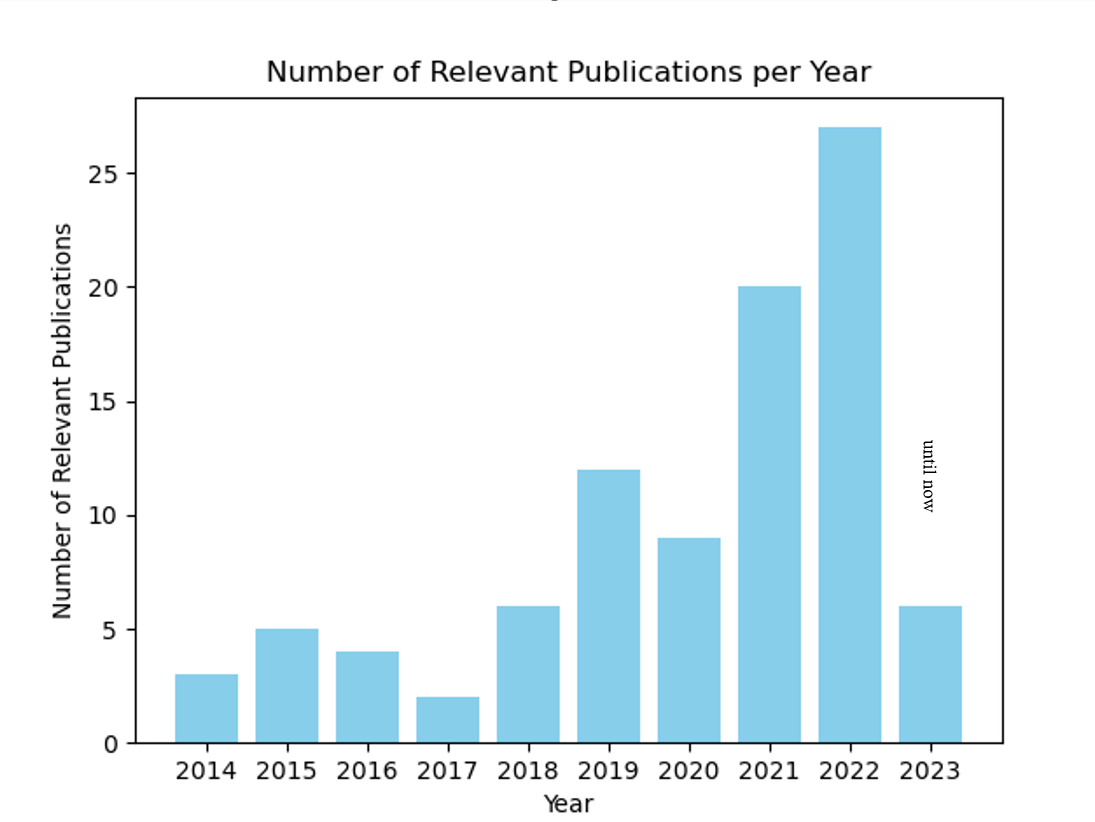}
\caption{Number of relevant publications per year.}
\label{fig:publications-per-year-chart}
\end{figure}

The chart shown in Figure~\ref{fig:publications-per-year-chart} depicts the number of relevant publications per year. The x-axis represents the years from 2014 to 2023, while the y-axis denotes the count of publications. The bars provide a clear visual representation of the distribution of relevant literature over the years, with a notable surge in publications from 2019 to 2022.
As illustrated in Figure~\ref{fig:publications-per-year-chart}, the majority of the relevant literature emerged within the last three years. This trend aligns with the rise in popularity of the topic, primarily driven by the introduction of pre-trained language models like BERT~\cite{subramani2020survey}.

Our literature collection methodology ensures the inclusion of recent breakthroughs and facilitates a comprehensive survey that captures the most current state of form understanding in noisy scanned documents. The selected papers will serve as the foundation for the subsequent sections of this survey, where we delve into various approaches, datasets, and comparisons to shed light on the advancements in this vibrant research area.

\section{Early Approaches in Document Understanding}
\label{sec:previous-work}

The evolution of document understanding witnessed the use of early models, predominantly based on rule-based algorithms. These methods can be broadly categorized into top-down and bottom-up approaches, each with its strengths and limitations~\cite{higashino1986knowledge, wang1994analysis, subramani2020survey}.

Top-down approaches, being efficient and effective, operate well when the document structure is fixed and well-defined. Typically, this method requires a high-level description of the logical structure of the document. However, its limitations become evident in scenarios with complex and varied document layouts. Therefore, top-down approaches are best suited for situations where document structures are known in advance.

On the other hand, bottom-up approaches are more time-consuming but offer the capability to analyze previously unknown document structures. These methods analyze the image to extract connected components, progressively merging them into coherent blocks (e.g., characters to words, words to lines, lines to paragraphs). The bottom-up strategy is more versatile, making it suitable for handling diverse document layouts, but it comes at the cost of increased computational overhead.

Form document processing, in particular, has seen the development of specialized methods based on form description~\cite[p.~1941]{tang1996automatic-document-processing-survey}. Researchers have explored form description languages (FDL) to represent the layout and logical structure of form documents using rectangular regions~\cite{higashino1986knowledge, tang1993documentanalysisbook, tang1996automatic-document-processing-survey}. FDL enables a comprehensive representation of the geometric and logical aspects of form documents, facilitating their automated processing.

For document understanding in general, earlier techniques included tree transformation to represent multi-article documents and the utilization of form definition languages~\cite[p.~1939]{tang1996automatic-document-processing-survey}. Tree transformations aimed to organize document content into a logical tree structure, aiding in content categorization and extraction. For instance, the separation of a document into header and body blocks involves labeling each block according to its type and then moving the blocks in the tree structure based on predefined rules.

These older models paved the way for document understanding, offering valuable insights and techniques that laid the foundation for subsequent research. However, the limitations of rule-based approaches and the demand for higher accuracy and adaptability led to the development of more sophisticated methods, including the recent advancements we explore in the subsequent sections of this survey.

\section{Multi-Modal Transformer Approaches}
\label{sec:approach-multi-modal}
In recent years, the trajectory of models in the document understanding domain has been characterized by a paradigm shift towards enriching transformer-based architectures with additional modalities beyond text. This evolution has led to breakthroughs in capturing layout, formatting, and contextual cues present in scanned documents. 

This section introduces a series of significant models that have emerged in recent years, contributing to the advancement of document understanding. We present these models chronologically, illustrating their innovative integration of multi-modal information and their unique contributions to the field. A comprehensive overview of the covered models, including the modalities they leverage and the types of approaches they embody, is provided in Table~\ref{tab:model-overview}.

\subsection{Layout-Visual Fusion Models}
Layout-Visual Fusion Models focuses on integrating both layout and visual information from document images with textual content to enhance document understanding. Document understanding is the task of automatically reading, analyzing, and extracting information from documents, such as forms, tables, invoices, etc\footnote{\url{https://github.com/microsoft/unilmhttps://learn.microsoft.com/en-us/microsoft-365/syntex/document-understanding-overview}}\footnote{\url{https://learn.microsoft.com/en-us/microsoft-365/syntex/document-understanding-overview}}. Layout-Visual Fusion Models combine features such as bounding box coordinates, image embeddings, and spatial relationships to capture the structural layout of documents, making them suitable for tasks that require precise spatial information, such as form understanding and table extraction~\cite{Shoubin03023,zheng2023layoutdiffusion,oral2022fusion}.

 \textbf{LayoutLM}~\cite{Xu2020LayoutLMPO} was the first transformer model that jointly trained text-level document information (using WordPiece~\cite{wu2016googles} representation) in conjunction with the text's 2D position representation.
    Additionally, layout information was added during fine-tuning using image-embeddings of token regions. It achieved new state-of-the-art results for several downstream tasks including form understanding. Like basic BERT models, it consists of a pre-training and a fine-tuning task. 
    For pre-training and fine-tuning, the text embedding of a word is used together with its two-dimensional position-embedding (i.e. the bounding box of the word on the document). The image embedding of the respective section on the document image is added during fine-tuning. 
    
    LayoutLM uses two pre-training tasks:
    \begin{itemize}
        \item \textbf{Masked Visual-Language Modeling (MVLM)} is inspired by the BERT task \emph{Masked Language Modeling (MLM)}, where random tokens of a sentence are masked using a special token and the model tries to predict the correct word for the gap.
        LayoutLM extends this idea by masking the word, but keeping its positional information for the prediction of the missing word.
        \item \textbf{Multi-label Document Classification} is an optional pre-training task, where the authors use document-labels from the IIT-CDIP dataset~\cite{Lewis2006BuildingAT} (discussed in Sec. 5) to guide the model to the correct document domain. Since this information is often not present for larger datasets, the authors propose that this task will be obsolete for future iterations.
    \end{itemize}
   
    The authors fine-tune the model using a form understanding, a receipt understanding, and a document image classification task.
    
\textbf{LayoutLMv2}~\cite{Xu2021LayoutLMv2MP} is the evolution of the LayoutLM model. The key idea is to further integrate the relation between text, layout, and visual information. The authors achieve this by incorporating image information during pre-training as well as the introduction of new pre-training tasks. \\
    The vanilla LayoutLM model relied on the traditional transformer self-attention mechanism, which has been replaced by the \textit{spatial aware self-attention mechanism}. This mechanism is build to extend the original attention \(\alpha_{ij}\) between the two tokens \(i\) and \(j\) by taking their 1D and 2D relations into account. 
    With \(\textbf{b}^{(1D)}\) being the learnable bias in one-dimensional space and \(\textbf{b}^{(2D_x)}\), \(\textbf{b}^{(2D_y)}\) denoting the 2D biases for x and y direction respectively, the new attention value \(\alpha_{ij}'\) is defined as 
    \[\alpha_{ij}' = \alpha_{ij} + \textbf{b}^{(1D)}_{j-i} + \textbf{b}^{(2D_x)}_{xj-xi} + \textbf{b}^{(2D_y)}_{yj-yi}.\]
    The output vector is then defined as the weighted average of the normalized spatial attention scores.
    
     For pre-training, LayoutLMv2 still uses the MVLM task of the original LayoutLM but adds two additional tasks. The first is \textit{text image alignment} (TIA), where image regions of random tokens are covered and the model needs to predict the labels of covered and not-covered tokens. \\
     The second addition is called \textit{text image matching} (TIM). Here, the model needs to predict whether text-information belongs to a given image. Negative examples for this task are acquired by replacing a document image with another one from the training set or by dropping it.
     
\subsection{Graph-Based Models}
Graph-Based Text Relationship Modeling is a technique that employs graph neural networks (GNNs) to capture intricate relationships between text segments in documents. GNNs can process data that can be represented as graphs, such as nodes and edges\footnote{\url{https://distill.pub/2021/gnn-intro/}}\footnote{\url{https://en.wikipedia.org/wiki/Graph_neural_network}}. These models use an encoder to process both textual and visual features, followed by a graph module that constructs a soft adjacency matrix to represent pairwise relationships between segments. This approach is particularly effective for capturing long-range dependencies and context in documents with complex textual interactions, such as relation extraction and entity linking~\cite{fu-etal-2019-graphrel,pham2023deep}.

\textbf{PICK}~\cite{yu2020pick} uses text and layout features in conjunction with a graph learning module to capture relationships between text segments. 
    The architecture of PICK\footnote{\url{https://github.com/wenwenyu/PICK-pytorch}} is built on three separate modules:
The \textit{encoder module} takes the original document as input. 
For each sentence of a document, the input vector consists of the text embeddings of characters, the bounding box, and the image embedding. Text segments are encoded using a transformer~\cite{vaswani2017attention} and image segments are encoded using a pre-trained ResNet CNN~\cite{he2015deep}. The text and image features are then combined by element-wise addition. 
    
The \textit{graph module} takes the combined representation of the encoder module and layout bounding boxes of the original document as input. It is trained to generate a \emph{soft adjacent matrix} that captures pairwise relationships between two graphs nodes.
The usage of graph learning allows this model to capture long-distance relationships between text segments, which is traditionally a problem for transformer-based language models.
    
The \textit{decoder module} that takes both the results of the encoder and the results of the graph module as input and forms the final output. The decoder uses a bidirectional long short-term memory network (BiLSTM) and a conditional random field algorithm (CRF) to perform sequence tagging on the union non-local sentence at a character level.

The \emph{Text Reading and Information Extraction} (\textbf{TRIE}) model~\cite{zhang2020trie} is unique in that the usually separated tasks of text-reading and text-understanding are fused into an end-to-end trained network. The general assumption is that both tasks can mutually benefit from each other during pre-training. The model consists of a \emph{text-reading module}, a \emph{multi-modal context block} and an \emph{information-extraction module}.  The \emph{text-reading module} uses a ResNet CNN~\cite{xie2017aggregated} in combination with a Feature Pyramid Network (FPN)~\cite{lin2017feature} to extract the convolutional features \( I \) for an input image. This image embedding is passed to a $detector$ function (usually anchor-based or segmentation-based approaches), which determines possible text regions \( B = (b_1, b_2, ..., b_m) \) as bounding boxes. The regions $B$ are fused with the image features $I$ using RoIAlign~\cite{he2018mask} to get the text-region (or visual) features \(C = (c_1, c_2, ..., c_m) \). Each $c_i$ is fed into an encoder/decode model to produce the corresponding characters $y = (y_1, y_2, ... y_T)$ where $T$ is the length of the label sequence.  These textual features are represented in textual feature vector $Z$ which is used together with visual feature vector $C$ for information extraction. The \emph{multi-modal context block} fuses textual, positional, and visual features created by the previous task. It makes use of a multi-head self-attention mechanism for the text embeddings, creating a text-context vector \( \Tilde{C} \). The \emph{information extraction module} uses the text-context \( \Tilde{C} \) and the visual context $C$ to separate entities and the text-embedding $Z$ to extract their entity values. First, the final context vector \( \overline{c_i}  \) is created as a weighted sum of the context vectors \( \Tilde{C} \) and $C$. This fused context vector is concatenated with the text-embedding, and the final feature vector is passed to bidirectional LSTM.

\subsection{Multi-Modal Fusion Models}
Multi-Modal Transformer Approaches are a type of large models that advance the fusion of modalities by introducing novel mechanisms to better integrate text, layout, and visual information. These models often propose modifications to the transformer architecture, such as spatial-aware self-attention mechanisms, to capture multi-dimensional relationships between tokens. They also introduce new pre-training tasks, like text-image matching and sequence length prediction, to enhance the understanding of document content and layout. These models are suitable for various multimodal tasks, such as document classification, image captioning, and video understanding~\cite{xu2023multimodal,feng2022multimodal,wang2023mutualformer}.

\textbf{SelfDoc}~\cite{li2021selfdoc} is a novel model in the way the textual features are encoded. 
    Instead of working on word-level embeddings, SelfDoc receives text grouped by semantically relevant components also referred to as \textit{document object proposals} (text-block, title, list, table, figure) as model input. 
    For the component-specific text extraction, the authors train a Faster R-CNN~\cite{ren2015faster} on public document datasets PubLayNet~\cite{zhong2019publaynet} and DocVQA~\cite{mathew2021docvqa} on a document object detection task. 
    The categorized regions are cropped from the document and passed to an OCR engine.

    The encoded visual and textual feature vectors are then passed separately through two distinct multi-head attention encoders. 
    Since those encoders do not share information between the visual and textual modality, they are referred to as \emph{single-modality encoders}.
    The resulting attention matrix of both modalities is then combined using a third encoder, making the system a \emph{cross-modality encoder}. This mechanism allows the SelfDoc model to use multi-modal document features during the pre-training phase.
    
\textbf{UniLMv2}~\cite{bao2020unilmv2} allows for pre-training of a unified language model consisting of an auto-encoding and and a partially auto-regressive model.
    The auto-encoding model predicts tokens by conditioning on context just like BERT~\cite{devlin2019bert}. 
    The partially auto-regressive model is based on masked- and pseudo-masked tokens. It can predict one or more masked (or pseudo-masked) tokens per factorization step. This approach allows the model to train on masked-token \emph{spans} instead of single tokens. The authors refer to this as \emph{Pseudo-Masked Language Modelling}.
    Masking of the data is done as follows: 15\% of original tokens are masked. 40\% of the time  n-gram blocks are masked and 60\% of the time single tokens are masked. The pre-training objective is to minimize the summed cross-entropy loss for the reconstruction of masked- and pseudo-masked-tokens in both models.
    The model's input consists of word token-embeddings, their absolute position embedding, and their segment embedding.     

\textbf{DocFormer}~\cite{appalaraju2021docformer}  improves on multi-modal training by fusing vision and language network layers. 
    The visual features are obtained by using a ResNet CNN~\cite{xie2017aggregated} at a lower-resolution embedding level. 
    The visual embedding is then reduced in dimensionality by applying a 1 x 1 convolution. 
    For the text embedding, DocFormer uses the OCR result from the document image and tokenizes it using WordPiece~\cite{wu2016googles}. The tokenized text $t_{tok}$ is in the form of $[CLS], t_1, t_2, ..., t_n$ where N = 511 and all further tokens  $t_i$ with $i > 511$ are ignored. The text embedding is padded using a [PAD] token that is ignored by the self-attention mechanism. The resulting text input feature is of the same shape as the visual feature and formally defined as
     $T=W_t(t_{tok})$
     where the weight $W_t$ is initialized from LayoutLM pre-training weights.
    
    For the spatial features the authors use the top-left and bottom-right coordinate of each word's bounding box in addition to the box's width $w$, height $h$, the euclidean distance from each corner of a box to each corner of the next box, and the centroid distance between boxes. 
    The latter two can be defined as $ A_{rel} = { A^{k+1}_{num} - A^k_{num}}; A \in (x,y); num \in (1,2,3,4,c) $ where $ num $ defines the direction of the relation ($c$ stands for the centroid coordinate). 
    Similar to many comparable models DocFormer also encodes a 1D position encoding $P^{abs}$ which defines the token sequence.
    
\textbf{StrucText}~\cite{li2021structext} model uses a multi-modal combination of visual and textual document features. 
    The textual embedding consists of a language token embedding where each text sentence is processed coherently to preserve semantic context. 
    This is different from other models where word tokens of a sentence are processed individually. A sentence S is a sorted set of words, with the line delimiters [CLS] and [SEP] inserted at the beginning (\(t_0\)) and the at the end (\(t_{n+1}\)) of a sentence respectively. 
    
    The final textual embedding is achieved by combining the language token embedding with the layout embedding (i.e. bounding box of token).
    
    For extractions of features of an image \(I\), ResNet \cite{xie2017aggregated} is used which creates the visual embedding. Additionally, a sequence id embedding is used to capitalize on text segment order.
    
    For pre-training, the authors used MVLM of LayoutLM in combination with two additional tasks.
    The first novel task is called \textit{sequence length prediction} (SLP) where the length of a text sequence has to be predicted based on the visual features of the document. 
    This task forces the model to learn from the image embedding.
    The second additional task is \textit{paired box direction} (PBD). 
    For two text segments \(i\) and \(j\), the model needs to predict in which direction segment \(j\) is located with respect to \(i\). 
    Here the 360-degree field is divided into eight buckets of identical size. 
    PBD is therefore a classification task that forces the model to learn spatial text segment relationships.

\begin{figure}[ht]
     \centering
     \includegraphics[width=.65\linewidth]{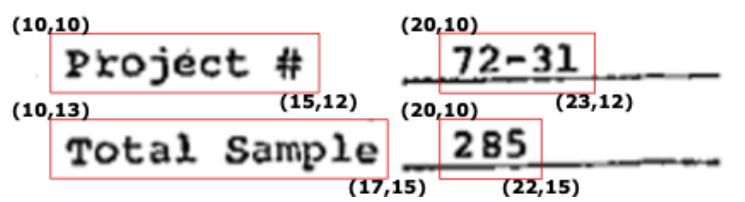}
     \includegraphics[width=.7\linewidth]{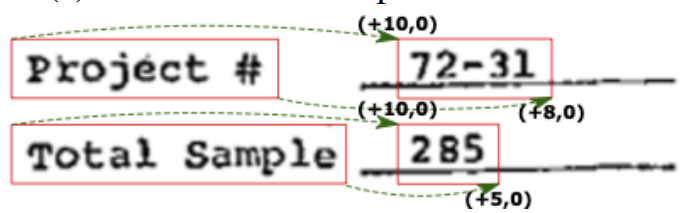}
     \caption{Absolute (top) vs. relative (bottom) 2D position representation \cite{hong2021bros}.}
     \label{fig:bros}
    \end{figure}

     \textbf{BROS}~\cite{hong2021bros} model does not incorporate the image of the text for pre-training, which is considered SOTA. 
    The authors claim that it is more effective to focus on the combination of text and position-information, ignoring the layout modality. 
    BROS introduces a custom text serializer that sorts the tokens in their respective reading order. 
    The model's features do not include absolute 2D positions of text, but a relative position representation based on the position of neighboring tokens as shown in Fig.~\ref{fig:bros}. This has the advantage that the model can capture relations between similar key-value pairs better compared to absolute spatial information. 
\subsection{Cross-Modal Interaction Models}
Cross-Modal Interaction Models are a type of artificial intelligence model that facilitate interactions between different modalities like text and layout in multimodal data. They employ attention mechanisms and fusion strategies to enable cross-modal information exchange and enhance the perception and understanding of the data. These models are suitable for various cross-modal tasks, such as document analysis, image-text retrieval, and video question answering.

\textbf{UDoc}~\cite{gu2023unified} is a framework that employs a multi-layer gated cross-attention encoder~\cite{lu2019vilbert,gu2020self}, enabling effective cross-modal interactions within a single model. Unlike traditional methods, UDoc takes into account the complex and diverse nature of documents by incorporating both textual and visual features. This approach addresses the challenges of hierarchical document structures, multimodal content, and spatial layout. UDoc employs three pretraining tasks: Masked Sentence Modeling (MSM)~\cite{ren2015faster}, Visual Contrastive Learning (VCL), and Vision-Language Alignment (VLA)~\cite{chen2020simple}. MSM predicts masked sentence embeddings, VCL focuses on quantized visual features, and VLA enforces alignment between textual and visual elements.
The document images are processed through feature extraction and quantization, providing a structured representation. The pretraining tasks, including MSM, VCL, and VLA, contribute to enhanced embeddings, capturing both local and global dependencies. 

\textbf{TILT}~\cite{powalski2021going} addresses the complex challenge of natural language comprehension in documents that extend beyond plain text by introducing the Text-Image-Layout Transformer (TILT) neural network architecture. TILT is designed to simultaneously capture layout information, visual features, and textual semantics, making it capable of handling documents with rich spatial layouts.  TILT model integrates layout representation through attention biases and contextualized visual information, leveraging pre-trained encoder-decoder Transformers~\cite{hewlett2016wikireading,raffel2020exploring}.
In the proposed TILT architecture, the model's strength lies in its ability to combine contextualized image embeddings~\cite{ethayarajh2019contextual} with semantic embeddings. By incorporating both layout and visual information, TILT enhances its understanding of documents, particularly those with complex structures like forms and tables. 

\subsection{Sequence-to-Sequence Models}
Encoder-Decoder and Sequence-to-Sequence Models involve encoders to process input data and decoders to generate sequential output. The encoders and decoders can be composed of different architectures, such as recurrent neural networks (RNNs), convolutional neural networks (CNNs), or transformers. They are used for tasks like text generation, translation, and sequence prediction, where the input and output have different lengths or structures~\cite{dalmia2019enforcing,michael2019evaluating,farina2023distillation,fu2023decoderonly}.

\textbf{GenDoc}~\cite{feng2023sequence} model employs an encoder-decoder architecture, allowing adaptability to diverse downstream tasks with varying output formats, which differentiates it from common encoder-only models. GenDoc's pre-training tasks encompass masked text prediction, masked image token prediction, and masked coordinate prediction, catering to multiple modalities. The model integrates modality-specific instructions, disentangled attention~\cite{peng2022ernie}, and a mixture-of-modality-experts(MoE)~\cite{bao2022vlmo} approach to effectively capture information from each modality.

The fusion of modalities within the encoder is achieved through disentangled attention, leading to improved layout effectiveness. The authors employ disentangled attentions for various interactions, including content-to-content, content-to-layout in the x-dimension, and content-to-layout in the y-dimension. This is represented by the following equations:

\[
\begin{aligned}
A_{ij}^{cc} &= \bm{Q}{i}^c{\bm{K}{j}^{c}}^{\intercal}, \\
A_{ij}^{cx} &= \bm{Q}{i}^c{\bm{K}{\delta_x(i, j)}^{x}}^{\intercal}, \\
A_{ij}^{cy} &= \bm{Q}{i}^c{\bm{K}{\delta_y(i, j)}^{y}}^{\intercal}, \\
A_{ij} &= A_{ij}^{cc} + A_{ij}^{cx} + A_{ij}^{cy}.
\end{aligned}
\]

Here, $A_{ij}$ represents the resulting attention score between positions $i$ and $j$;
$A_{ij}^{cc}$ is the conventional attention score from content query $\boldsymbol{Q}^c$ to content key $\boldsymbol{K}^c$ for positions $i$ and $j$;
$A_{ij}^{cx}$ and $A_{ij}^{cy}$ denote the disentangled attention scores from content to layout;
$\delta_{\ast}(i,j)$ denotes the relative distance function between positions $i$ and $j$.

In contrast, modality separation within the decoder employs the MoE strategy, which activates task-specific experts for natural language, image token prediction, and coordinate prediction tasks. The pre-training tasks include text infilling, masked image token prediction, and masked coordinate prediction, employing a shared vocabulary.

\begin{figure}[t]
\centering
\begin{minipage}[b]{0.45\linewidth}
\includegraphics[width=\linewidth]{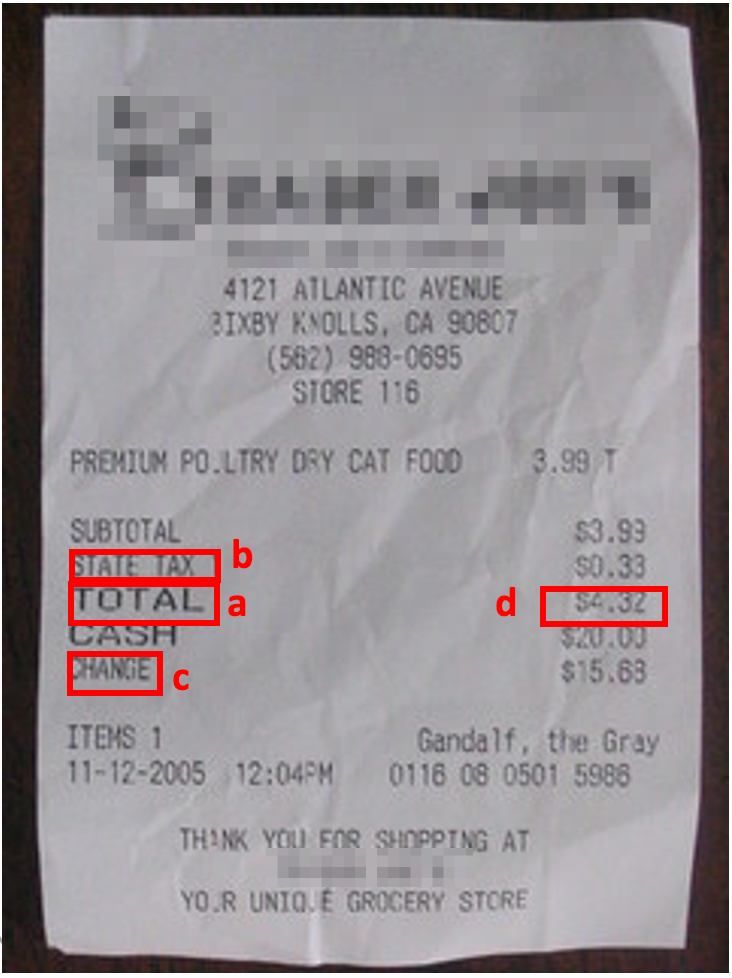}
\caption{\textbf{Token-to-Line}~\cite{appalaraju2023docformerv2}}
\label{fig:token_to_line}
\end{minipage}
\hspace{0.05\linewidth}
\begin{minipage}[b]{0.45\linewidth}
\includegraphics[width=\linewidth]{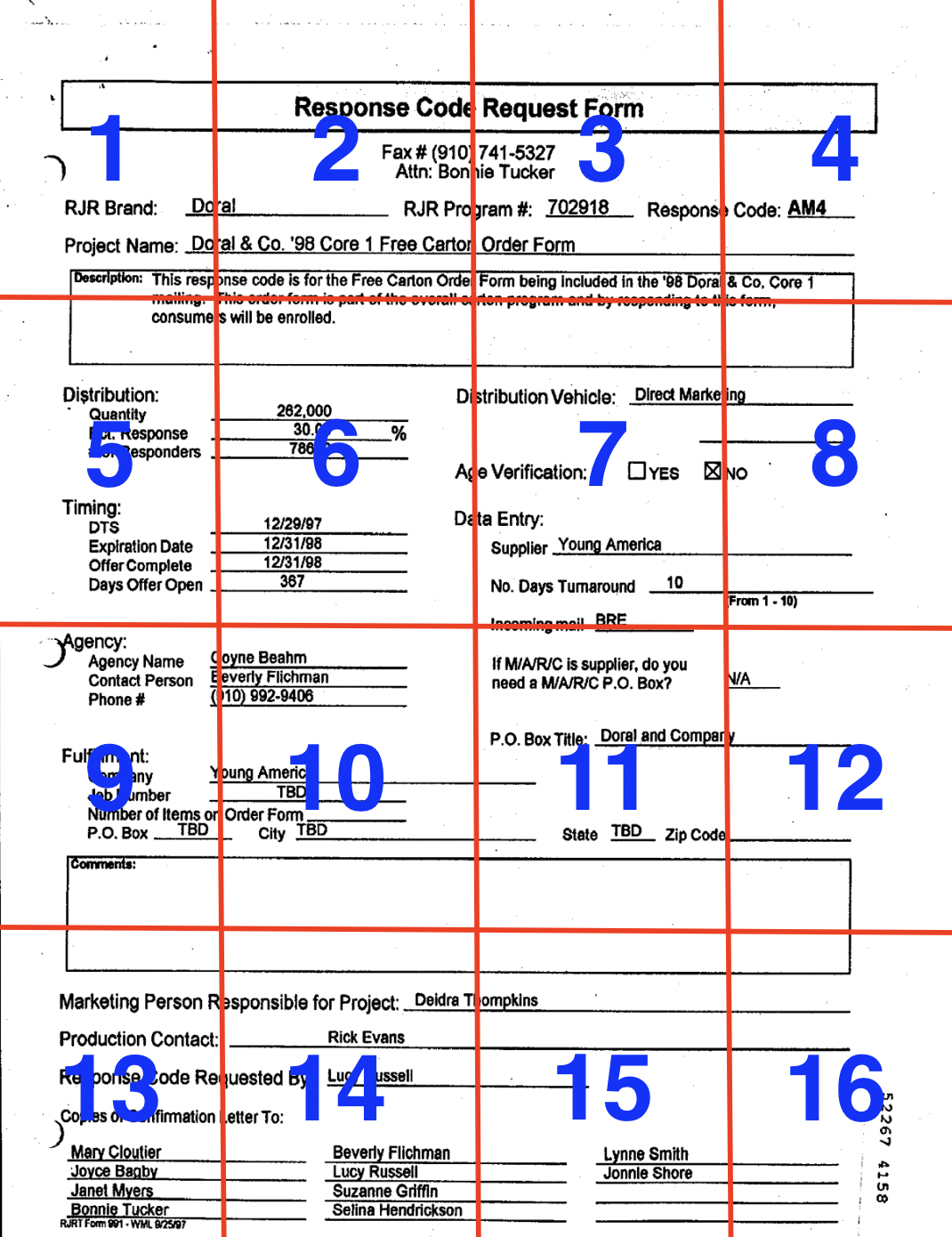}
\caption{\textbf{Token-to-Grid} ~\cite{appalaraju2023docformerv2}}
\label{fig:token_to_grid}
\end{minipage}
\end{figure}

\textbf{DocFormerv2}~\cite{appalaraju2023docformerv2} comprises a multi-modal encoder~\cite{chen2020transformer} tailored to process both textual and spatial characteristics inherent in documents. The model adopts traditional transformer-based language modeling to handle textual content while introducing inventive pre-training tasks like token-to-line and token-to-grid predictions to capture document layout and structure as shown in Fig~\ref{fig:token_to_line} and~\ref{fig:token_to_grid}. This approach enables the model to understand the spatial arrangement of document elements.

To effectively incorporate spatial information from document images, the authors introduce spatial features that encode critical elements such as position, size, and category. By encoding these elements, DocFormerv2 gains an understanding of the document's structure and the relationships between different components.

A key strength of DocFormerv2 lies in its adept multi-modal fusion techniques~\cite{li2022two}. By employing self-attention mechanisms, the model seamlessly integrates textual and spatial features, enabling it to comprehensively understand document content and layout. This fusion facilitates improved performance in a range of document understanding tasks.

\textbf{FormNet}~\cite{lee2022formnet} is designed to improve the extraction of information from complex form-like documents. These documents, which often contain tables, columns, and intricate layouts, pose challenges for traditional sequence models. FormNet addresses this issue by incorporating structural information into the sequence modeling process. FormNet contains two components: Rich Attention and Super-Tokens. Rich Attention enhances attention score calculation by considering both semantic relationships and spatial distances between tokens, enabling the model to capture the underlying layout patterns more effectively. Super-Tokens are constructed for each word in the form through graph convolutions, recovering local syntactic relationships that may have been lost during serialization. The information extraction problem is formulated as sequential tagging for tokenized words, with the goal of predicting key entity classes for each token. The authors build upon the Extended Transformer Construction (ETC) framework, which efficiently handles long sequences. ETC employs sparse global-local attention, and FormNet enhances positional encoding through Rich Attention.

\subsection{Layout Representation and Language-Independent Models}
Layout Representation Models~\cite{hsiao2019flat2layout,tang2023unifying,wang2022lilt} focus on capturing and representing the spatial arrangement of components in documents. They encode layout information for tasks like structure analysis and extraction. Language-Independent Models~\cite{voutharoja2023language,gao2023enabling,nogueira2020navigation} designed to work across multiple languages. They often utilize pre-training in one language and fine-tuning in others, enabling document understanding in various languages.

\textbf{MCLR}~\cite{Shi_2023_WACV}  proposed method employs a cell-based layout representation. This representation utilizes row and column indexes to define the position of components within a document, aligning with human reading habits and facilitating a more intuitive understanding of layout structures.   The MCLR architecture comprises three main components: the cell-based layout, the multi-scale layout, and data augmentation.

In the cell-based layout, the traditional challenge of learning spatial relationships from coordinates is addressed by using row and column indexes. This enables the model to learn whether components share the same row or column and the number of components in each row or column. The approach also provides a more human-centered representation, aligning with natural reading patterns and enhancing the detection of latent rules within the document layout. Furthermore, the authors introduce a multi-scale layout representation by employing word- and token-level cells.

\textbf{LiLT}~\cite{wang2022lilt} introduces an innovative solution to a critical challenge in structured document understanding (SDU).  The authors propose the Language-independent Layout Transformer (LiLT), a novel framework that enables pre-training on documents from a single language and subsequent fine-tuning for use with multiple languages.

The core architecture of LiLT is founded upon a parallel dual-stream Transformer model that efficiently processes both text and layout information. During pre-training, LiLT decouples the text and layout aspects of the document and employs a bi-directional attention complementation mechanism (BiACM) to facilitate interaction between these modalities. The model is then fine-tuned with a focus on downstream tasks using off-the-shelf pre-trained textual models. This process ensures that LiLT learns and generalizes the layout knowledge from monolingual documents to multilingual contexts. 

\subsection{Hybrid Transformer Architectures}
Hybrid Transformer Architectures~\cite{li-etal-2021-structurallm,yu2023structextv2} leverage innovative designs to address the challenges of computational complexity and diverse structural representations in document understanding. These models often incorporate novel attention mechanisms, such as Symmetry Cross-Attention (SCA)~\cite{zhai2023fast}, and utilize hourglass transformer architectures with merging and extension blocks to efficiently process multi-modal information. These designs enhance both performance and efficiency in analyzing complex documents.

\textbf{StructuralLM}~\cite{li-etal-2021-structurallm}  is a successor to the LayoutLM model.
	It differs from LayoutLM by operating on 2D cell position embeddings instead of 2D word embeddings. 
	The core idea is to better capture word relations by grouping them. 
	Each document \(d\) is represented by a sequence of cells \( d = \{ c_1 , ... , c_n \} \) and each cell is represented by a sequence of words \(c_i = \{ w^1_i , ... , c^m_i \} \), where each word \( w_i \) of cell \(c_i\) has the same 2D position embedding. 
	In addition to this, the 2D embedding receives 1D position information per token (i.e. the sequence number of the token) to leverage word order within a cell, and the embedding of the word itself using WordPiece~\cite{wu2016googles}.
	
	The first pre-training task for the StructuralLM model is the MVLM task introduced by LayoutLM \cite{Xu2020LayoutLMPO}. 
	Since positional information is not dropped during MVLM and words of the same cell share common position information, StructuralLM can leverage the cell grouping during this task.

	The second pre-training task is called \emph{cell position classification} (CPC). 
	Here, a set of scanned documents is split into \(N\) areas. For a random cell of the documents, the model needs to predict where the cell belongs using its center coordinate. 
	This is in essence a multinomial classification task where each label \( [1,N] \) belongs to a document section.
 
\textbf{StrucTexTv2}~\cite{yu2023structextv2} a model that combines visual and textual information to effectively analyze document images. The proposed approach employs a unique pre-training strategy with two self-supervised tasks: Mask Language Modeling (MLM)~\cite{devlin2018bert} and Mask Image Modeling (MIM)~\cite{bao2021beit,chen2022context}, where text region-level masking is utilized to predict both visual and textual content. The core architecture of StrucTexTv2 comprises two main components: a visual extractor (CNN) and a semantic module (Transformer). The visual extractor processes the input document image using convolutional layers, extracting visual features from different stages of downsampling. The semantic module employs the Transformer architecture, which takes the flattened visual features and transforms them into a 1D sequence of patch token embeddings. A relative position embedding is added to the token embeddings, and the standard transformer processes these embeddings to generate enhanced semantic features. The resulting feature maps are reshaped to 2D context feature maps and up-sampled with a Feature Pyramid Network (FPN) strategy~\cite{lin2017feature}.

\textbf{Fast-StrucTexT}~\cite{zhai2023fast} is an innovative and efficient transformer-based framework designed for document understanding tasks. The proposed model addresses the challenges posed by the computational complexity of traditional transformers while efficiently handling the diverse structural representations within documents. Fast-StrucTexT incorporates a novel hourglass transformer architecture, modality-guided dynamic token merging, and Symmetry Cross-Attention (SCA)~\cite{wang2022rtformer} to achieve both high performance and efficiency. 

The SCA module functions as a two-way cross-attention mechanism that facilitates interaction between different modes. In this module, one set of model features serves as the query, while another set of features from a different mode serves as both the key and value. This arrangement enables the calculation of cross-attention in both visual and textual modalities.
The SCA is defined by the following equations:

1. Cross-Attention (CA) between feature sets \(f_n\) and \(f_m\) is given by:
\[
\begin{aligned}
& \text{CA}(f_n|f_m) = W_o \cdot \sigma\left(\frac{F_q(f_n)F^T_k(f_m)}{\sqrt{d}}\right) \cdot F_v(f_m),
\end{aligned}
\]
where \(F_q\), \(F_k\), and \(F_v\) are linear projections for query, key, and value respectively. \(\sigma\) represents the softmax function, \(d\) is the hidden size, and \(W_o\) is the output weight.

2. SCA between feature sets \(f_n\) and \(f_m\) is defined as:
\[
\begin{aligned}
& \text{SCA}(f_n, f_m) = \{\text{CA}(f_n|f_m), \text{CA}(f_m|f_n)\}.
\end{aligned}
\]

Fast-StrucTexT leverages an hourglass transformer design, integrating Merging-Blocks and Extension-Blocks to efficiently process multi-modal information and handle complex document structures. The Merging-Blocks aggregate tokens to capture higher-level context, enhancing efficiency without compromising semantic richness. Complementarily, Extension-Blocks recover merged tokens, ensuring fine-grained details are preserved. The modality-guided dynamic token merging mechanism selectively merges tokens based on relevance, enabling diverse structural representations.

\begin{table}[]
\small
\centering
\caption{An overview of the covered models in terms of their used modalities.}
\label{tab:model-overview}

\def\arraystretch{1.1}
\begin{adjustbox}{max width=0.5\textwidth}
\begin{tabular}{l|cccc|l}
\textbf{} & \multicolumn{4}{c|}{\textbf{Modalities}} &  \\ \hline
Model & \multicolumn{1}{c|}{\rotatebox{90}{Text }} & \multicolumn{1}{c|}{\rotatebox{90}{1D Pos. }} & \multicolumn{1}{c|}{\rotatebox{90}{2D Pos. }} & \rotatebox{90}{Image } & Approach type \\ \hline
BERT & \multicolumn{1}{c|}{\checkmark} & \multicolumn{1}{c|}{\checkmark} & \multicolumn{1}{c|}{} &  &  \\
LayoutLM & \multicolumn{1}{c|}{\checkmark} & \multicolumn{1}{c|}{\checkmark} & \multicolumn{1}{c|}{\checkmark} & \checkmark & image-based fine-tuning \\
UniLMv2 & \multicolumn{1}{c|}{\checkmark} & \multicolumn{1}{c|}{\checkmark} & \multicolumn{1}{c|}{\checkmark} &  & seq2seq \\
PICK & \multicolumn{1}{c|}{\checkmark} & \multicolumn{1}{c|}{\checkmark} & \multicolumn{1}{c|}{\checkmark} & \checkmark & graph-learning \\
TRIE & \multicolumn{1}{c|}{\checkmark} & \multicolumn{1}{c|}{\checkmark} & \multicolumn{1}{c|}{\checkmark} & \checkmark & multi-task (OCR + IE) \\
LayoutLMv2 & \multicolumn{1}{c|}{\checkmark} & \multicolumn{1}{c|}{\checkmark} & \multicolumn{1}{c|}{\checkmark} & \checkmark & spatial-aware self-attention \\
StructuralLM & \multicolumn{1}{c|}{\checkmark} & \multicolumn{1}{c|}{\checkmark} & \multicolumn{1}{c|}{\checkmark} & \checkmark & text-segment grouping \\
SelfDoc & \multicolumn{1}{c|}{\checkmark} & \multicolumn{1}{c|}{\checkmark} & \multicolumn{1}{c|}{\checkmark} & \checkmark & text-segment grouping \\
DocFormer & \multicolumn{1}{c|}{\checkmark} & \multicolumn{1}{c|}{\checkmark} & \multicolumn{1}{c|}{\checkmark} & \checkmark & discrete multi-modal \\
StrucText & \multicolumn{1}{c|}{\checkmark} & \multicolumn{1}{c|}{\checkmark} & \multicolumn{1}{c|}{\checkmark} & \checkmark & text-segment grouping \\
BROS & \multicolumn{1}{c|}{\checkmark} & \multicolumn{1}{c|}{\checkmark} & \multicolumn{1}{c|}{\checkmark} &  & using relative 2D positions \\

StrucTexTv2 & \multicolumn{1}{c|}{\checkmark} & \multicolumn{1}{c|}{\checkmark} & \multicolumn{1}{c|}{\checkmark} & \multicolumn{1}{c|}{\checkmark} & multi-modal fusion \\
Fast-StrucTexT & \multicolumn{1}{c|}{\checkmark} & \multicolumn{1}{c|}{\checkmark} & \multicolumn{1}{c|}{\checkmark} & \multicolumn{1}{c|}{\checkmark} & efficient transformer \\
DocFormerv2 & \multicolumn{1}{c|}{\checkmark} & \multicolumn{1}{c|}{\checkmark} & \multicolumn{1}{c|}{\checkmark} & \multicolumn{1}{c|}{\checkmark} & multi-modal encoder \\
UDoc & \multicolumn{1}{c|}{\checkmark} &\multicolumn{1}{c|}{\checkmark} &\multicolumn{1}{c|}{\checkmark} & \multicolumn{1}{c|}{\checkmark} & gated cross-attention \\
TILT & \multicolumn{1}{c|}{\checkmark} & \multicolumn{1}{c|}{\checkmark} & \multicolumn{1}{c|}{\checkmark} & \multicolumn{1}{c|}{\checkmark} & Text-Image-Layout Transformer \\
GenDoc & \multicolumn{1}{c|}{\checkmark} & \multicolumn{1}{c|}{\checkmark} & \multicolumn{1}{c|}{\checkmark} &\multicolumn{1}{c|}{\checkmark} & modality fusion, disentangled attention \\
MCLR & \multicolumn{1}{c|}{\checkmark} & \multicolumn{1}{c|}{\checkmark} & \multicolumn{1}{c|}{\checkmark} & \multicolumn{1}{c|}{\checkmark} & cell-based layout, multi-scale layout \\
LiLT & \multicolumn{1}{c|}{\checkmark} & \multicolumn{1}{c|}{\checkmark} & \multicolumn{1}{c|}{\checkmark} & \multicolumn{1}{c|}{\checkmark} & language-independent layout \\
\end{tabular}
\end{adjustbox}
\end{table}



\section{Datasets}
\label{sec:datasets}
In this section, we introduce commonly used datasets for form understanding tasks.
For a comparison of basic features of the datasets, see Tab.~\ref{tab:dataset-comparison}.

\begin{table*}[t]
\centering
\caption{Comparison between individual datasets. All listed datasets are publicly available.}
\label{tab:dataset-comparison}
\begin{adjustbox}{width=1\textwidth}
\begin{tabular}{lccccc}
\hline
\textbf{Dataset} & \textbf{Document Content} & \textbf{Color} & \textbf{Used Languages} & \textbf{\#Documents} & \textbf{Document Type} \\ \hline
FUNSD & Machine Written / Handwritten & B/W & EN & 199 & Form Documents \\ 
XFUND & Machine Written / Handwritten & B/W & Multi-lingual & 199/language & Form Documents \\ 
NAF & Machine written pre-printed text and mostly handwritten input text & B/W & EN & 865 & Multiple \\ 
IIT-CDIP & Machine Written, Handwritten & B/W & EN & $6,919,192$ & Multiple \\ 
RVL-CDIP & Machine Written, Handwritten & B/W & EN & $400,000$  & Multiple \\ 
PubLayNet & Machine Written & Color & EN & $364,232$ & Medical Literature \\ 
SROIE & Machine Written with occasional handwriting & Color, B/W & EN & $1,000$ & Receipts \\ 
CORD & Machine Written with occasional handwriting & Color & EN & $1,000$ released, $11,000$ total & Receipts \\ 
DocVQA & Machine Written, Handwritten & Color, B/W & EN & $12,767$ & Multiple \\ 
Form-NLU & Machine Written, Handwritten & Color, B/W & EN & $10,857$ & Multiple \\ 
VRDU & Machine Written & Color, B/W & EN & $2,556$ & Ad-buy  \\ 

\hline

\end{tabular}
\end{adjustbox}
\end{table*}

\textbf{RVL-CDIP and IIT-CDIP:} The \textit{Ryerson Vision Lab Complex Document Information Processing} RVL-CDIP dataset~\cite{harley2015icdar} was created to evaluate Deep CNNs\footnote{\url{https://www.cs.cmu.edu/~aharley/rvl-cdip/}}.
The dataset is a subset of the \textit{IIT Complex Document Information Processing Test Collection} IIT-CDIP~\cite{Lewis2006BuildingAT,cdip2021ftpdata} dataset, which itself contains a collection of documents from the Legacy Tobacco Documents Library (LTDL) created by the University of California San Francisco~\cite{schmidt2002building}.
IIT-CDIP consists of $6,919,192$ document records in XML format, each of which contains text recognized by OCR and metadata information (document names, geographical information, etc.) of various structure and quality.
Additionally, IIT-CDIP contains source images of scanned documents~\cite{lewis:iit-cdip-readme:2021}.

In total, the RVL-CDIP dataset consists of $400,000$ grayscale images separated into 16 classes (including form documents) with $25,000$ images per class. 
The dataset is split into 320k training, 40k validation, and 40k test images. 


\textbf{FUNSD:}
The dataset for Form Understanding in Noisy Scanned Documents (FUNSD)~\cite{jaume2019} is a collection of 199 fully annotated forms from various fields like marketing, science, and more\footnote{\url{https://guillaumejaume.github.io/FUNSD/}}. 
All forms are rasterized, low-resolution one-page forms containing noise. 
More specifically, the authors manually selected $3,200$ eligible documents from the RVL-CDIP dataset~\cite{Lewis2006BuildingAT} form category and used random sampling to gain a final set of $199$ documents. 
Most of the textual content is machine-written, which might not reflect real-life scenarios where handwritten content can be more prominent.

All annotations used for text detection were done by external contractors. 
The remaining tasks were annotated using an undisclosed annotation tool. 
Every form in the dataset is encoded in a JSON file where each form is considered as a list of interlinked \textit{semantic entities} (i.e. a name field that is interlinked with its corresponding answer box). 
Each semantic entity contains an identifier, a label, a bounding box, a list of links with other entities, and a list of words.
In total, the dataset contains $9,707$ semantic entities and $5,304$ relations.


The authors of \cite{vu2020revising} have found several errors and inconsistencies in the FUNSD dataset including incorrect textual contents and bounding boxes as well as inconsistencies in annotating relations. They released an updated version of the dataset that fixes the discovered issues\footnote{\url{https://drive.google.com/drive/folders/1HjJyoKqAh-pvtg3eQAmrbfzPccQZ48rz?fbclid=IwAR2ouj5Sh0vkcKAMNfSoZjSM7vSpGnbK-AowWZZ8_Lltcn34hr7_nVcazu0}}. 

\textbf{XFUND}~\cite{xu2021layoutxlm} is an extension to FUNSD where the dataset is translated into seven other languages, namely Chinese, Japanese, Spanish, French, Italian, German, and Portuguese.
Like FUNSD, the dataset contains information on both semantic entities in a document and their relations to one another\footnote{\url{https://github.com/doc-analysis/XFUND}}.
The dataset was manually created by gathering form templates in the above-mentioned languages from the internet. 
Each form is filled out exactly once with synthetic information to prevent repetition of forms and avoid sharing sensitive information. 
Forms are filled out either digitally or with handwriting and then scanned to again be similar to the original FUNSD dataset.

To generate bounding boxes for each semantic entity, the authors use the Microsoft Read API~\cite{microsoft2022readapi}. 
Relation extraction is then done manually by annotators.
In total, the XFUND dataset contains $199$ forms per language ($1,393$ total) split into training/test splits of $149$/$50$ documents respectively.

\textbf{NAF}~\cite{davis2019deep,davis2019NAFgithub} is an annotated dataset created from historical form images of the United States National Archives. 
The forms have varied layouts and are noisy from degradation and machinery that was used to print them. 
A form in the dataset typically contains English pre-printed text and English input text consisting of handwriting, typed text, and stamped content.
In total, the dataset contains $865$ annotated grayscale form images.
It is undisclosed how they were annotated.

The NAF dataset is manually split into a training, a validation, and a test set such that each set contains images with distinct form layouts. 
With this, the authors create a scenario where a classifier trained with the training set will not have seen the form layouts of the test set.

\textbf{PubLayNet}~\cite{zhong2019publaynet} is a dataset that was \textit{automatically} created by matching the XML representations and the content of $1,162,856$ scientific PDF articles from PubMed Central\textsuperscript{TM}. 
The dataset was created primarily to advance research in \textit{document layout analysis}, meaning it is not specifically made for form detection\footnote{\url{https://github.com/ibm-aur-nlp/PubLayNet}}. 
The annotations were created by parsing the PDF representation of the article using the \texttt{PDFMiner}~\cite{shinyama2019pdfminer} software and then matching the generated layout with the respective XML representation. 

While not mentioned in the paper, most of the 
annotated documents seem to have been removed during the quality control procedure as only a total of $364,232$ documents remain in the dataset after this step. 
The dataset is split into $340,391$ pages for training, $11,858$ pages for development, and $11,983$ pages for testing.
Both development and testing sets have been selected from multiple different journals and contain a sufficient number of tables, lists, and figures.

The \textbf{SROIE}~\cite{Huang_2019} dataset was created as part of the ICDAR 2019 competition on \textit{Scanned Receipt OCR and Information Extraction}\footnote{\url{https://drive.google.com/drive/folders/1ShItNWXyiY1tFDM5W02bceHuJjyeeJl2}}. 
The dataset contains 1000 scanned receipt images with different annotations depending on one of three competition tasks (\textit{Text Localisation}, \textit{OCR}, and \textit{Key-Information Extraction}). 
Receipts in the dataset may have poor paper, ink, or print quality, be of low resolution, and contain scanning artifacts. 
Further, receipts may contain unneeded interfering texts in complex layouts, long texts, and small font sizes. 
Each receipt contains English character text fields like goods name and digit fields like unit price and total cost.
The paper does not mention the origin of the receipts. 



\textbf{CORD: }
The \textit{Consolidated Receipt Dataset} for post-OCR parsing~\cite{park2019cord} was created to be the first public dataset containing box-level text and parsing class annotations.
Parsing class annotations are represented as a combination of one of eight superclasses (i.e. \textit{store}, \textit{payment}, and \textit{menu}) and their respective subclasses (i.e. \textit{name}, \textit{address}, and \textit{price});
e.g., 
\textit{menu}.\textit{price}.

The dataset itself contains $11,000$ Indonesian receipts from shops and restaurants gathered through crowdsourcing.
Images were manually annotated using a web-based annotation tool where each image is first annotated and then checked for correctness and compliance with the annotation guidelines.
Sensitive information like credit card numbers or a person's full name have been blurred in the final set of receipts.
Out of the $11,000$ receipts, only $1,000$ have been made available to the public due to concerns of accidentally publishing sensitive personal data.
According to one of the authors of the paper, future data release is unclear~\cite{parker:github-cord-release:2021}.



\textbf{DocVQA:}~\cite{mathew2021docvqa,mathew2021docvqadataset} is a dataset of $12,767$ document images of varied types for the task of \textit{Visual Question Answering}.
The documents are sourced from documents of the UCSF Industry Documents Library~\cite{ucsflibrary:IndustryDocumentsLibrary}. 
Individual documents were hand-picked to provide suitable image quality and tables, forms, lists, and figures over long-running text fields.
The documents are split randomly in an 80/10/10 ratio of train/validation/test splits respectively.

A total of $50,000$ questions and answers have been defined on the documents by remote workers using a three-step web-based annotation tool.
In the first step, a worker defines up to ten question-answer pairs for a document. 
In the second step, another worker will try to answer the defined questions without seeing the pre-defined answers.
If no answers of the first step match with any of the answers of the second step, the question is moved to a third step where the authors of the paper can manually edit the question-answer pairs.

Besides question-answer pairs, the dataset also contains one of nine question types per question. 
An example of a question type is \textit{table/list} which specifies if table or list understanding is required to answer the question.
A question can be assigned more than one question type.

\textbf{Form-NLU:}~\cite{ding2023form} is a comprehensive dataset designed to advance the field of Natural Language Understanding (NLU) in the context of form-based documents. The dataset comprises a total of $857$ document images, 6k form keys and values, and 4k table keys and values. These documents encompass a variety of form types, including digital, printed, and handwritten. Form-NLU constitutes a portion of the publicly accessible financial form dataset originating from Form 604, which was amassed by SIRCA\footnote{\url{https://asic.gov.au/regulatory-resources/forms/forms-folder/604-notice-of-changeof-
interests-of-substantial-holder/}}. It provides text records of substantial shareholder notifications filed with the Australian Stock Exchange (ASX)\footnote{\url{https://www2.asx.com.au/}} from 2003 to 2015. The authors segmented the annotation into three specialized sub-tasks: Key annotation for 12 types of Keys, Value annotation for paired Values, and Non-Key-Value annotation for other components like Title, Section, and Others. Each annotator handled one sub-task.



\textbf{VRDU:} is a benchmark dataset for Visually-rich Document Understanding, consisting of two distinct datasets: Ad-buy Forms, comprising 641 documents, and Registration Forms, comprising 1,915 documents. These datasets are designed to address challenges in extracting structured data from visually complex documents. The benchmark incorporates five key desiderata: rich schema, layout-rich documents, diverse templates, high-quality OCR results, and token-level annotation. 
The benchmark focuses on three tasks: 
\begin{enumerate}
    \item Single Template Learning (STL): the model is trained and evaluated on documents that belong to a single template. This means that the training, validation, and testing sets all contain documents that share the same structural layout or template. The goal of this task is to assess the model's ability to extract structured data when presented with a consistent and familiar document layout.
    \item  Mixed Template Learning (MTL): the model is trained and evaluated on documents that come from a set of templates. The training, validation, and testing sets include documents from different but predefined templates. This task evaluates the model's capacity to generalize its learning to diverse document layouts and templates that it has encountered during training.
    \item Unseen Template Learning (UTL): The Unseen Template Learning task challenges the model to generalize beyond its training experience. In this task, the model is trained on documents from a subset of templates and is then evaluated on documents with templates it has never encountered during training. The goal is to assess the model's ability to adapt to new, previously unseen document layouts and templates.
\end{enumerate}

The evaluation toolkit includes a type-aware fuzzy matching algorithm to assess extraction performance. Repeated, unrepeated, and hierarchical entity names are included, with post-processing heuristics to handle unrepeated and hierarchical entities. 
\section{Experiments Results}
In this section, we provide a comprehensive comparison of selected models on various datasets, including FUNSD, SROIE, CORD, RVL-CDIP, DocVQA, and Publaynet. We evaluate the models' performance using a range of metrics, including Precision, Recall, F1 Score, Accuracy, Average Normalized Line Similarity (Anls), and Mean Average Precision (MAP).
\subsection{Evaluation Metrics}

We utilize several evaluation metrics to assess the effectiveness of the models in extracting structured information from scanned documents. The key evaluation metrics are as follows:

\subsubsection{Precision, Recall, F1 Score, and Accuracy}

Precision, Recall, F1 Score, and Accuracy are standard metrics used to evaluate the performance of models in classification tasks. These metrics are calculated as follows:

\begin{equation}
\text{Precision} = \frac{\text{True Positives}}{\text{True Positives} + \text{False Positives}}
\end{equation}

\begin{equation}
\text{Recall} = \frac{\text{True Positives}}{\text{True Positives} + \text{False Negatives}}
\end{equation}

\begin{equation}
\text{F1 Score} = \frac{2 \times \text{Precision} \times \text{Recall}}{\text{Precision} + \text{Recall}}
\end{equation}

\begin{equation}
\text{Accuracy} = \frac{\text{True Positives} + \text{True Negatives}}{\text{Total Samples}}
\end{equation}

Where:
\begin{itemize}
    \item \text{True Positives} (TP) are the correctly predicted positive samples.
    \item \text{False Positives} (FP) are the incorrectly predicted positive samples.
    \item \text{False Negatives} (FN) are the incorrectly predicted negative samples.
    \item \text{True Negatives} (TN) are the correctly predicted negative samples.
    \item \text{Total Samples} is the total number of samples in the dataset.
\end{itemize}

\subsubsection{ Average Normalized Levenshtein Similarity (Anls)}
Anls~\cite{biten2019scene} captures the OCR mistakes applying a slight penalization in case of correct intended responses, but badly recognized. It also makes use of a threshold of value 0.5 that dictates whether the output of the metric will be the ANLS if its value is equal to or bigger than 0.5 or 0 otherwise. The key point of this threshold is to determine if the answer has been correctly selected but not properly recognized, or on the contrary, the output is a wrong text selected from the options and given as an answer.

More formally, the ANLS between the net output and the ground truth answers is given by the following equation. Where $N$ is the total number of questions, $M$ total number of GT answers per question, $a_{ij}$ the ground truth answers where $i = \{0, ..., N\}$, and $j = \{0, ..., M\}$, and $o_{qi}$ be the network's answer for the ith question $q_i$:

\begin{equation}
\mathrm{ANLS} = \frac{1}{N} \sum_{i=0}^{N} \left(\max_{j} s(a_{ij}, o_{qi}) \right)
\end{equation}

Where:
where $s(\cdot, \cdot)$ is defined as follows:

\begin{equation}
    s(a_{ij}, o_{qi}) = \begin{cases}
    1 - \mathrm{NL}(a_{ij}, o_{qi}), & \text{if } \mathrm{NL}(a_{ij}, o_{qi}) < \tau \\
    0,                               & \text{if } \mathrm{NL}(a_{ij}, o_{qi}) \geq \tau
    \end{cases}
\end{equation}

\subsubsection{Mean Average Precision (MAP)}

MAP is a metric commonly used to evaluate the quality of ranked lists in information retrieval tasks. It involves calculating the Average Precision (AP) for each query and then averaging these values across all queries. MAP is also used in object detection to evaluate the performance of a model. In object detection, the goal is to identify and localize objects in an image or video. The MAP metric takes into account both the accuracy of the detections and the precision of the detections. The formula for calculating MAP is as follows:

\begin{equation}
\text{MAP} = \frac{1}{Q} \sum_{q=1}^{Q} \text{AP}(q)
\end{equation}

Where:
\begin{itemize}
    \item \(Q\) is the total number of queries.
    \item \(\text{AP}(q)\) is the Average Precision for query \(q\), calculated as the average precision of relevant documents at each position in the ranked list:
\end{itemize}

\begin{equation}
\text{AP}(q) = \frac{1}{N_q} \sum_{k=1}^{N_q} \text{Precision}(k) \times \text{Relevance}(k)
\end{equation}

Where:
\begin{itemize}
    \item \(N_q\) is the total number of retrieved documents for query \(q\).
    \item \(\text{Precision}(k)\) is the precision at position \(k\) in the ranked list.
    \item \(\text{Relevance}(k)\) is an indicator function that takes the value 1 if the document at position \(k\) is relevant to query \(q\), and 0 otherwise.
\end{itemize}

\subsection{Model Comparison}
\label{sec:comparison}
In this section, We present a comparison of selected models on the FUNSD, SROIE, CORD, RVL-CDIP, DocVQA, and Publaynet datasets in Table~\ref{tab:model-FUNSDcomparison},~\ref{tab:model-SROIEcomparison},~\ref{tab:model-Publaynetcomparison},~\ref{tab:model-DocVQAcomparison},~\ref{tab:model-RVL-CDIPcomparison},and~\ref{tab:model-CORDcomparison}. The table provides the Precision, Recall, F1 Score, Accuracy, Anls, MAP, and the number of parameters for each model on each dataset.

\subsubsection{FUNSD Dataset}
\begin{table}[h]
\center
\scriptsize

\caption{Performance Comparison of Document Layout Analysis Models on the FUNSD Dataset.}
\begin{adjustbox}{width=0.5\textwidth}
\begin{tabular}{lccccc}
\hline
\multicolumn{1}{c}{\textbf{Model}} & \textbf{Dataset} & \textbf{Precision} & \textbf{Recall} & \textbf{F1} &\textbf{\#Params} \\ \hline
LayoutLM\textsubscript{Base} & FUNSD & 0.7597 & 0.8155 & 0.7866 & 113M \\
LayoutLM\textsubscript{Large} & FUNSD & 0.7596 & 0.8219 & 0.7895 & 343M \\
LayoutLMv2\textsubscript{Base} & FUNSD & 0.8029 & 0.8539 & 0.8276 & 200M \\

BROS\textsubscript{Base} & FUNSD & 0.8116 & 0.8502 & 0.8305 & 110M \\
StrucText & FUNSD & 0.8568 & 0.8097 & 0.8309 & 107M \\
DocFormer\textsubscript{Base} & FUNSD & 0.8076 & 0.8609 & 0.8334 & 183M \\
SelfDoc & FUNSD & - & - & 0.8336 & -  \\
LayoutLMv2\textsubscript{Large} & FUNSD & 0.8324 & 0.8519 & 0.8420 & 426M \\

BROS\textsubscript{Large} & FUNSD & 0.8281 & 0.8631 & 0.8452 & 340M \\
DocFormer\textsubscript{Large} & FUNSD & 0.8229 & 0.8694 & 0.8455 & 536M \\
FormNet& FUNSD & 0.8521 & 0.8418 & 0.8469 &217M \\
StructuralLM\textsubscript{Large} & FUNSD & 0.8352 & 0.8681 & 0.8514 & 355M \\ 
UDoc& FUNSD & - & - & 0.8720 & 274M \\
DocFormerv2\textsubscript{Base}& FUNSD & 0.8915 & 0.876 & 0.8837 & 232M \\
DocFormerv2\textsubscript{Large}& FUNSD & 0.8988 & 0.8792 & 0.8889 & 750M \\
StrucTexTv2\textsubscript{Small} & FUNSD & - & - & 0.8923 & 107M \\
Fast-StrucTexT\textsubscript{Linear} & FUNSD & - & - & 0.8950 & - \\
LayoutLMv3\textsubscript{Base}& FUNSD & - & - & 0.9029 & 133M \\
Fast-StrucTexT\textsubscript{ResNet-18} & FUNSD & - & - & 0.9035 & - \\
StrucTexTv2\textsubscript{Large} & FUNSD & - & - & 0.9182 & 238M \\
LayoutLMv3\textsubscript{Large}& FUNSD & - & - & 0.9208 & 368M \\

MCLR\textsubscript{Large}& FUNSD & - & - & 0.9352 &368M \\
MCLR\textsubscript{Base}& FUNSD & - & - & 0.9376 & 133M \\

DiT\textsubscript{Base} & FUNSD  & 0.9470 & 0.9307 & 0.9388 & 87M \\
DiT\textsubscript{Large} & FUNSD  & \textbf{0.9452} & \textbf{0.9336} & \textbf{0.9393} & 304M \\

\hline
\end{tabular}
\end{adjustbox}
\label{tab:model-FUNSDcomparison}
\end{table}

The FUNSD dataset, renowned for form understanding tasks, serves as a crucial benchmark for evaluating various models' performance. A comprehensive analysis of the comparison results presented in Table~\ref{tab:model-FUNSDcomparison} yields valuable insights into the models' abilities. Notably, the transition from LayoutLM to LayoutLMv2 highlights the impact of introducing spatial-aware self-attention, leading to improved F1 scores. Noteworthy performance is observed from LayoutLMv3 \textit{BASE}, which outperforms its predecessors with an impressive F1 score of 0.9029. StructuralLM \textit{LARGE} stands out with an F1 score of 0.8514, indicating the success of its approach. StrucTexTv2 \textit{LARGE}, despite its lower parameter count, achieves a competitive F1 score of 0.9182, demonstrating its efficiency in form understanding tasks. Additionally, DiT \textit{BASE} and DiT \textit{LARGE} showcase promising results with F1 scores of 0.9388 and 0.9393, respectively. 

\subsubsection{SROIE Dataset}
\begin{table}[h]
\center
\scriptsize
\caption{Performance Comparison of Document Layout Analysis Models on SROIE Dataset.}
\begin{adjustbox}{width=0.5\textwidth}
\begin{tabular}{lcccccccc}
\hline
\multicolumn{1}{c}{\textbf{Model}} & \textbf{Dataset} & \textbf{Precision} & \textbf{Recall} & \textbf{F1} & \textbf{\#Params} \\ \hline
LayoutLM\textsubscript{Base} & SROIE  & 0.9438 & 0.9438 & 0.9438  &113M \\
LayoutLM\textsubscript{Large} & SROIE  & 0.9524 & 0.9524 & 0.9524 &343M \\
PICK & SROIE  & 0.9679 & 0.9546 & 0.9612 &- \\
TRIE & SROIE  & - & - & 0.9618  &- \\ 
LayoutLMv2\textsubscript{Base} & SROIE  & 0.9625 & 0.9625 & 0.9625  &200M \\
LayoutLMv2\textsubscript{Large} & SROIE  & 0.9661 & 0.9661 & 0.9661 &426M \\
StrucText & SROIE  & 0.9584 & 0.9852 & 0.9688  &107M \\
Fast-StrucTexT\textsubscript{Linear} & SROIE & - & - & 0.9712 & - \\
Fast-StrucTexT\textsubscript{ResNet-18} & SROIE & - & - & 0.9755 & - \\
TILT\textsubscript{Base}& SROIE  & - & - & 0.9765  &230M \\
TILT\textsubscript{Large}& SROIE  & - & - & \textbf{0.9810} &780M \\
\hline
\end{tabular}
\end{adjustbox}
\label{tab:model-SROIEcomparison}
\end{table}

The evaluation of various models on the SROIE dataset, which is centered around receipt understanding and key-information extraction tasks, provides insights into the capabilities of these models for form-based understanding. Table~\ref{tab:model-SROIEcomparison} presents a comprehensive comparison of the performance achieved by different models on this dataset.

LayoutLMv2 \textit{BASE} and LayoutLMv2 \textit{LARGE} demonstrate their prowess in the receipt understanding task. Both models achieve high precision, recall, and F1 scores of 0.9625 and 0.9661 respectively. StrucText exhibits strong recall with a score of 0.9852, showcasing its effectiveness in capturing key information from receipts. Its F1 score of 0.9688 indicates a balanced trade-off between precision and recall, making it a viable option for the SROIE dataset.

\subsubsection{Publaynet Dataset}
\begin{table}[h]
\center
\scriptsize
\caption{Performance Comparison of Document Layout Analysis Models on Publaynet Dataset.}

\begin{tabular}{lcccccccc}
\hline
\multicolumn{1}{c}{\textbf{Model}} & \textbf{Dataset} & \textbf{MAP}&\textbf{\#Params} \\ \hline
StrucTexTv2\textsubscript{224*224}& Publaynet  &0.8590&-\\
StrucTexTv2\textsubscript{512*512}& Publaynet  &0.9210&-\\
UDoc& Publaynet  &0.9390&-\\
GenDoc\textsubscript{Base}& Publaynet  &0.9440&- \\
StrucTexTv2\textsubscript{960*960}& Publaynet  &0.9490&-\\
DiT\textsubscript{Base} & Publaynet & 0.9450 &87M \\
DiT\textsubscript{Large} & Publaynet & 0.9490 &304M   \\
LayoutLMv3\textsubscript{Base}& Publaynet  &\textbf{0.9510}&133M\\

\hline
\end{tabular}
\label{tab:model-Publaynetcomparison}
\end{table}

The PubLayNet dataset, despite its primary focus on advancing document layout analysis, has garnered attention from researchers in the field of form understanding due to its extensive annotations and diverse scientific content. As showcased in Table~\ref{tab:model-Publaynetcomparison}, various models have been evaluated on the PubLayNet dataset. Among the evaluated models, LayoutLMv3 \textit{Base} stands out with a high Mean Average Precision (MAP) of 0.9510. This suggests that the model's spatial-aware self-attention mechanism and architecture contribute to effectively capturing layout information, even in the context of intricate scientific documents. Similarly, DiT \textit{LARGE} showcases promising results with a MAP of 0.9490, demonstrating its capability in processing diverse document layouts for form understanding. GenDoc \textit{Base}, while not specifically designed for layout analysis, still achieves a competitive MAP of 0.9440, indicating its generalizability to document understanding tasks. UDoc also demonstrates its effectiveness in this domain, achieving a MAP of 0.9390. StrucTexTv2 models, operating at different resolutions (224x224, 512x512, and 960x960), exhibit varying levels of performance, with the 960x960 variant achieving the highest MAP of 0.9490.

\subsubsection{DocVQA Dataset}

\begin{table}[h]
\center
\scriptsize
\caption{Performance Comparison of Document Layout Analysis Models on DocVQA Dataset.}
\begin{tabular}{lcccccccc}
\hline
\multicolumn{1}{c}{\textbf{Model}} & \textbf{Dataset} &\textbf{Anls}&\textbf{\#Params} \\ \hline
LayoutLMv2\textsubscript{Large}& DocVQA  & 0.7808  &200M \\
LayoutLMv3\textsubscript{Large}& DocVQA  & 0.7876  &133M \\
GenDoc\textsubscript{Base}& DocVQA  & 0.7883 &- \\
LayoutLMv3\textsubscript{Large}& DocVQA  & 0.8337  &368M \\
TILT\textsubscript{Base}& DocVQA  & 0.8392  &230M \\
GenDoc\textsubscript{Large}& DocVQA  & 0.8449  &- \\
StructuralLM\textsubscript{Large} & DocVQA & 0.8610  & 355M \\ 
LayoutLMv2\textsubscript{Large}& DocVQA  & 0.8672  &426M \\
TILT\textsubscript{Large}& DocVQA & 0.8705  &780M \\
DocFormerv2\textsubscript{Base}& DocVQA  & 0.8720 & 232M \\
DocFormerv2\textsubscript{Large}& DocVQA  & \textbf{0.8784} & 750M \\

\hline
\end{tabular}
\label{tab:model-DocVQAcomparison}
\end{table}

The evaluation of various models on the DocVQA dataset provides insights into their effectiveness in addressing the challenges posed by document-based visual question answering. Table~\ref{tab:model-DocVQAcomparison} presents a comprehensive comparison of the performance achieved by different models on this dataset. StructuralLM \textit{LARGE} stands out as a strong performer, boasting an analysis score of $0.8610$. This achievement underscores the model's ability to integrate structural information from documents and effectively answer questions based on visual and textual cues. Both LayoutLMv3 \textit{Large} and LayoutLMv2 \textit{Large} demonstrate their prowess in document understanding, achieving scores of $0.7876$ and $0.8672$ respectively. These models leverage advanced self-attention mechanisms to capture contextual and spatial relationships in documents, enabling them to tackle complex questions that involve both textual and visual components.

\subsubsection{RVL-CDIP Dataset}
\begin{table}[h]
\center
\scriptsize
\caption{Performance Comparison of Document Layout Analysis Models on RVL-CDIP Dataset.}

\begin{tabular}{lcccccccc}
\hline
\multicolumn{1}{c}{\textbf{Model}} & \textbf{Dataset} &  \textbf{Accuracy} &\textbf{\#Params} \\ \hline

MCLR\textsubscript{base}& RVL-CDIP  & 0.9070  &133M \\
BEiT\textsubscript{Base} & RVL-CDIP  & 0.9109 &87M \\
DiT\textsubscript{Base}\cite{li2022dit} & RVL-CDIP & 0.9211 &87M \\
DiT\textsubscript{Large}\cite{li2022dit} & RVL-CDIP & 0.9269 &304M   \\
StrucTexTv2\textsubscript{Small} & RVL-CDIP  & 0.9340  &28M \\
UDoc&  RVL-CDIP  & 0.9364 &274M \\
GenDoc\textsubscript{Base}& RVL-CDIP  & 0.9380  &- \\
SelfDoc & RVL-CDIP &0.9381 & -  \\

LayoutLM\textsubscript{Large}  & RVL-CDIP & 0.9442 &160M \\
LayoutLM\textsubscript{Large}  & RVL-CDIP &  0.9443 &390M \\

GenDoc\textsubscript{Large}& RVL-CDIP  & 0.9450  &- \\
StrucTexTv2\textsubscript{Large} & RVL-CDIP  & 0.9462 &238M \\ 

TILT\textsubscript{Base}& RVL-CDIP  & 0.9525  &230M \\
LayoutLMv2\textsubscript{Base} & RVL-CDIP  & 0.9525 &200M \\
LayoutLMv3\textsubscript{Base} & RVL-CDIP  & 0.9544 &133M \\
DocFormer\textsubscript{Large} & RVL-CDIP  & 0.9550 &536M \\
TILT\textsubscript{Large}& RVL-CDIP & 0.9552  &780M \\
LayoutLMv2\textsubscript{Large} & RVL-CDIP & 0.9564 &426M \\
LayoutLMv3\textsubscript{Large} & RVL-CDIP & 0.9593 &368M \\

StructuralLM\textsubscript{Large} & RVL-CDIP  & 0.9608 &355M \\ 
DocFormer\textsubscript{Base} & RVL-CDIP  & \textbf{0.9617} & 183M \\
\hline
\end{tabular}
\label{tab:model-RVL-CDIPcomparison}

\end{table}
The RVL-CDIP dataset offers a distinct dimension in the realm of form understanding, featuring multi-class single-label classification tasks. The comprehensive model comparison presented in Table~\ref{tab:model-RVL-CDIPcomparison} provides valuable insights into various models' performances. Among these models, the DocFormer architecture emerges as a standout performer, particularly DocFormer \textit{Base}, which attains an exceptional accuracy score of 0.9617, positioning itself at the forefront. This accomplishment underscores the efficacy of the DocFormer approach in tackling the challenges posed by multi-class classification tasks within the form understanding domain.

A noteworthy observation arises from the juxtaposition of DocFormer \textit{LARGE} and DocFormer \textit{BASE}, accentuating the interplay between model complexity and performance enhancement. Although DocFormer \textit{LARGE} achieves marginally lower accuracy than its \textit{BASE} counterpart, it does so with a notable increase in the number of model parameters. This trade-off underlines the nuanced relationship between model size and the achieved accuracy, showcasing the balance that must be struck when considering practical deployment and computational efficiency.

Moreover, this dataset comparison affirms the competitive capabilities of other models, such as StructuralLM \textit{LARGE}, LayoutLMv3, LayoutLMv2, and StrucTexTv2 \textit{LARGE}, each contributing to the diverse landscape of form understanding. Furthermore, the inclusion of various baseline models like BEiT and DiT provides a comprehensive perspective on the state-of-the-art advancements in multi-class classification on the RVL-CDIP dataset.
\subsubsection{CORD Dataset}
\begin{table}[h]
\center
\scriptsize
\caption{Performance Comparison of Document Layout Analysis Models on CORD Dataset.}

\begin{adjustbox}{width=0.5\textwidth}
\begin{tabular}{lccccc}
\hline
\multicolumn{1}{c}{\textbf{Model}} & \textbf{Dataset} & \textbf{Precision} & \textbf{Recall} & \textbf{F1} &\textbf{\#Params} \\ \hline
LayoutLM\textsubscript{Base} & CORD & 0.9437 & 0.9508 & 0.9472  &113M \\
LayoutLM\textsubscript{Large} & CORD & 0.9432 & 0.9554 & 0.9493 & 343M \\
LayoutLMv2\textsubscript{Base} & CORD & 0.9453 & 0.9539 & 0.9495 & 200M \\

TILT\textsubscript{Base}& CORD & - & - & 0.9511 & 230M \\
LayoutLMv2\textsubscript{Large} & CORD & 0.9565 & 0.9637 & 0.9601 &426M \\ 
DocFormer\textsubscript{Base} & CORD & 0.9652 & 0.9614 & 0.9633  & 183M \\
TILT\textsubscript{Large}& CORD & - & - & 0.9633 & 780M \\
LayoutLMv3\textsubscript{Base}& CORD & - & - & 0.9656 & 133M \\
GenDoc\textsubscript{Base}& CORD & - & - & 0.9659 & -  \\
Fast-StrucTexT\textsubscript{Linear} & CORD & - & - & 0.9665 &- \\
DocFormerv2\textsubscript{Base}& CORD & 0.9751 & 0.9610 & 0.9680 & 232M \\
GenDoc\textsubscript{Large}& CORD & - & - & 0.9697 & -  \\
DocFormer\textsubscript{Large} & CORD & 0.9725 & 0.9674 & 0.9699  &536M \\

Fast-StrucTexT\textsubscript{ResNet-18} & CORD & - & - & 0.9715  &- \\
MCLR\textsubscript{Base}& CORD & - & - & 0.9723 & 133M \\
FormNet& CORD & 0.9802 & 0.9655 & 0.9728 & 345M \\
LayoutLMv3\textsubscript{Large}& CORD & - & - & 0.9746 & 368M \\
MCLR\textsubscript{Base}& CORD & - & - & 0.9749 &368M \\
DocFormerv2\textsubscript{Large}& CORD & 0.9771 & 0.9770 & 0.9770 & 750M \\

UDoc\textsubscript{Large}& CORD & - & - & \textbf{0.9813} & 274M \\
\hline
\end{tabular}
\end{adjustbox}
\label{tab:model-CORDcomparison}
\end{table}
While the CORD dataset is more oriented towards receipt understanding, The models demonstrate a range of strengths in capturing different aspects of the documents, including textual content, spatial arrangement, and overall structure. A comprehensive analysis of the comparison results presented in Table~\ref{tab:model-CORDcomparison}. Notably, the LayoutLM variants,  LayoutLM \textit{BASE} and LayoutLM \textit{LARGE} exhibit strong performance with F1 scores of 0.9472 and 0.9493, respectively, benefiting from their layout-aware self-attention mechanisms. DocFormer models, particularly DocFormer \textit{BASE} and DocFormer \textit{LARGE}, excel in capturing complex structural patterns and linguistic context, achieving impressive F1 scores of 0.9633 and 0.9699. The LayoutLMv2 models also perform well, with LayoutLMv2 \textit{LARGE} achieving an F1 score of 0.9601. The Fast-StrucTexT models, including Fast-StrucTexT \textit{Linear} and Fast-StrucTexT \textit{ResNet-18}, demonstrate solid performance with F1 scores of 0.9665 and 0.9715, respectively, indicating the efficiency of their hourglass transformer architecture. UDoc \textit{LARGE} emerges as a top performer, achieving an impressive F1 score of 0.9813, showcasing the effectiveness of its gated cross-attention mechanism in capturing both textual and spatial information. Models like TILT \textit{Base} and GenDoc \textit{Base} also show competitive performance, highlighting their potential in addressing document understanding challenges. MCLR \textit{base} and MCLR \textit{base} achieve strong F1 scores of 0.9723 and 0.9749, respectively, emphasizing the success of their cell-based layout representation. FormNet achieves a high accuracy of 0.9728, showcasing its proficiency in extracting information from complex form-like documents.

\subsubsection{Overall Insights}

We explored different datasets to understand various forms, along with comparisons of the models used, which have provided insights into the current state of this field. Through this analysis, we were able to uncover both opportunities and challenges that extend beyond individual instances, shedding light on the broader landscape of form understanding. A key observation is the rich diversity of approaches employed to tackle the form understanding tasks, as revealed by model comparisons across different datasets. Notably, the consistent performance of LayoutLMv2 models highlights a wide range of spatial-aware self-attention mechanisms. One standout during these comparisons is the diversity of using DocFormer models across various datasets. Their consistent ability to deliver competitive results underscores their potential to adapt and excel across of form understanding tasks. This adaptability positions the DocFormer models as a strong contender for addressing a range of document-related challenges.

The balance between model complexity, quantified by the number of parameters, and the resulting performance gains. While larger models tend to yield slightly improved results, the considerable increase in resource requirements highlights the need for thoughtful evaluation of the trade-offs between computational demands and performance enhancements when selecting models for specific tasks. The distinct focus of each dataset opens up avenues for the development of specialized models. For example, the SROIE dataset, centered around receipts, showcases how models like TILT \textit{Large} can excel in context-specific scenarios. This underscores the potential for efficient form understanding even in domain-specific contexts. The real-world datasets introduced challenges inherent to complex and unstructured data, such as noise, layout variations, and diverse document formats. These hurdles underscore the necessity for models that can robustly handle such complexities, emphasizing the importance of building solutions that are both resilient and adaptable.

\section{Conclusions}
\label{sec:conclusion}

In this paper, we provided an extensive overview of modern approaches for form understanding. We showed that current state-of-the-art models are all based on transformer models, which makes them easy to use and adapt to a variety of tasks due to the pre-training / fine-tuning approach. MVLM is used by 6 models (LayoutLM, UniLMv2, LayoutLMv2, StructuralLM, StrucText, BROS). All of the presented models used multiple document modalities. The textual baseline transformer is always extended with text-position information on either word, sentence, or text-segment level. Especially the relative position encoding used by the BROS model seems promising. The layout information, which is provided by extracting text-image features, also seems crucial for good performance, as shown by the performance increase of LayoutLMv2 compared to its predecessor. 

\bibliography{emnlp2023}

\begin{thebibliography}{100}

\bibitem{davis2019deep}
Brian Davis, Bryan Morse, Scott Cohen, Brian Price, and Chris Tensmeyer.
\newblock Deep visual template-free form parsing, 2019.

\bibitem{abdallah2022tncr}
Abdelrahman Abdallah, Alexander Berendeyev, Islam Nuradin, and Daniyar Nurseitov.
\newblock Tncr: Table net detection and classification dataset.
\newblock {\em Neurocomputing}, 473:79--97, 2022.

\bibitem{zhong2019publaynet}
Xu~Zhong, Jianbin Tang, and Antonio~Jimeno Yepes.
\newblock Publaynet: largest dataset ever for document layout analysis, 2019.

\bibitem{Lewis2006BuildingAT}
David~D. Lewis, Gady Agam, Shlomo~Engelson Argamon, Ophir Frieder, David~A. Grossman, and Jefferson Heard.
\newblock Building a test collection for complex document information processing.
\newblock {\em Proceedings of the 29th annual international ACM SIGIR conference on Research and development in information retrieval}, 2006.

\bibitem{nurseitov2021handwritten}
Daniyar Nurseitov, Kairat Bostanbekov, Daniyar Kurmankhojayev, Anel Alimova, Abdelrahman Abdallah, and Rassul Tolegenov.
\newblock Handwritten kazakh and russian (hkr) database for text recognition.
\newblock {\em Multimedia Tools and Applications}, 80:33075--33097, 2021.

\bibitem{nurseitov2021classification}
Daniyar Nurseitov, Kairat Bostanbekov, Maksat Kanatov, Anel Alimova, Abdelrahman Abdallah, and Galymzhan Abdimanap.
\newblock Classification of handwritten names of cities and handwritten text recognition using various deep learning models.
\newblock {\em arXiv preprint arXiv:2102.04816}, 2021.

\bibitem{kim2022ocrfree}
Geewook Kim, Teakgyu Hong, Moonbin Yim, Jeongyeon Nam, Jinyoung Park, Jinyeong Yim, Wonseok Hwang, Sangdoo Yun, Dongyoon Han, and Seunghyun Park.
\newblock Ocr-free document understanding transformer, 2022.

\bibitem{toiganbayeva2022kohtd}
Nazgul Toiganbayeva, Mahmoud Kasem, Galymzhan Abdimanap, Kairat Bostanbekov, Abdelrahman Abdallah, Anel Alimova, and Daniyar Nurseitov.
\newblock Kohtd: Kazakh offline handwritten text dataset.
\newblock {\em Signal Processing: Image Communication}, 108:116827, 2022.

\bibitem{kasem2022deep}
Mahmoud Kasem, Abdelrahman Abdallah, Alexander Berendeyev, Ebrahem Elkady, Mahmoud Abdalla, Mohamed Mahmoud, Mohamed Hamada, Daniyar Nurseitov, and Islam Taj-Eddin.
\newblock Deep learning for table detection and structure recognition: A survey.
\newblock {\em arXiv preprint arXiv:2211.08469}, 2022.

\bibitem{irie2019language}
Kazuki Irie, Albert Zeyer, Ralf Schl{\"u}ter, and Hermann Ney.
\newblock Language modeling with deep transformers.
\newblock {\em arXiv preprint arXiv:1905.04226}, 2019.

\bibitem{wang2019language}
Chenguang Wang, Mu~Li, and Alexander~J. Smola.
\newblock Language models with transformers, 2019.

\bibitem{abdallah2023amurd}
Abdelrahman Abdallah, Mahmoud Abdalla, Mohamed Elkasaby, Yasser Elbendary, and Adam Jatowt.
\newblock Amurd: annotated multilingual receipts dataset for cross-lingual key information extraction and classification.
\newblock {\em arXiv preprint arXiv:2309.09800}, 2023.

\bibitem{9716741}
Kai Han, Yunhe Wang, Hanting Chen, Xinghao Chen, Jianyuan Guo, Zhenhua Liu, Yehui Tang, An~Xiao, Chunjing Xu, Yixing Xu, Zhaohui Yang, Yiman Zhang, and Dacheng Tao.
\newblock A survey on vision transformer.
\newblock {\em IEEE Transactions on Pattern Analysis and Machine Intelligence}, 45(1):87--110, 2023.

\bibitem{wu2021cvt}
Haiping Wu, Bin Xiao, Noel Codella, Mengchen Liu, Xiyang Dai, Lu~Yuan, and Lei Zhang.
\newblock Cvt: Introducing convolutions to vision transformers, 2021.

\bibitem{li2021supervision}
Yangguang Li, Feng Liang, Lichen Zhao, Yufeng Cui, Wanli Ouyang, Jing Shao, Fengwei Yu, and Junjie Yan.
\newblock Supervision exists everywhere: A data efficient contrastive language-image pre-training paradigm.
\newblock {\em arXiv preprint arXiv:2110.05208}, 2021.

\bibitem{devlin2018bert}
Jacob Devlin, Ming-Wei Chang, Kenton Lee, and Kristina Toutanova.
\newblock Bert: Pre-training of deep bidirectional transformers for language understanding.
\newblock {\em arXiv preprint arXiv:1810.04805}, 2018.

\bibitem{radford2019language}
Alec Radford, Jeff Wu, Rewon Child, David Luan, Dario Amodei, and Ilya Sutskever.
\newblock Language models are unsupervised multitask learners.
\newblock 2019.

\bibitem{abdallah2023generator}
Abdelrahman Abdallah and Adam Jatowt.
\newblock Generator-retriever-generator: A novel approach to open-domain question answering.
\newblock {\em arXiv preprint arXiv:2307.11278}, 2023.

\bibitem{abdallah2023exploring}
Abdelrahman Abdallah, Bhawna Piryani, and Adam Jatowt.
\newblock Exploring the state of the art in legal qa systems.
\newblock {\em arXiv preprint arXiv:2304.06623}, 2023.

\bibitem{han2022survey}
Kai Han, Yunhe Wang, Hanting Chen, Xinghao Chen, Jianyuan Guo, Zhenhua Liu, Yehui Tang, An~Xiao, Chunjing Xu, Yixing Xu, et~al.
\newblock A survey on vision transformer.
\newblock {\em IEEE transactions on pattern analysis and machine intelligence}, 45(1):87--110, 2022.

\bibitem{arnab2021vivit}
Anurag Arnab, Mostafa Dehghani, Georg Heigold, Chen Sun, Mario Lu{\v{c}}i{\'c}, and Cordelia Schmid.
\newblock Vivit: A video vision transformer.
\newblock In {\em Proceedings of the IEEE/CVF international conference on computer vision}, pages 6836--6846, 2021.

\bibitem{Xu2020LayoutLMPO}
Yiheng Xu, Minghao Li, Lei Cui, Shaohan Huang, Furu Wei, and Ming Zhou.
\newblock Layoutlm: Pre-training of text and layout for document image understanding.
\newblock {\em Proceedings of the 26th ACM SIGKDD International Conference on Knowledge Discovery \& Data Mining}, 2020.

\bibitem{binmakhashen2019document}
Galal~M Binmakhashen and Sabri~A Mahmoud.
\newblock Document layout analysis: A comprehensive survey.
\newblock {\em ACM Computing Surveys (CSUR)}, 52(6):1--36, 2019.

\bibitem{subramani2020survey}
Nishant Subramani, Alexandre Matton, Malcolm Greaves, and Adrian Lam.
\newblock A survey of deep learning approaches for ocr and document understanding.
\newblock {\em arXiv preprint arXiv:2011.13534}, 2020.

\bibitem{igorevna2022document}
Andreeva~Elena Igorevna, Bulatov~Konstantin Bulatovich, Nikolaev~Dmitry Petrovich, Petrova~Olga Olegovna, Savelev~Boris Igorevich, and Slavin~Oleg Anatolevich.
\newblock Document image analysis and recognition: a survey.
\newblock {\em computer optics}, 46(4):567--589, 2022.

\bibitem{liang2005camera}
Jian Liang, David Doermann, and Huiping Li.
\newblock Camera-based analysis of text and documents: a survey.
\newblock {\em International Journal of Document Analysis and Recognition (IJDAR)}, 7:84--104, 2005.

\bibitem{shigarov2023table}
Alexey Shigarov.
\newblock Table understanding: Problem overview.
\newblock {\em Wiley Interdisciplinary Reviews: Data Mining and Knowledge Discovery}, 13(1):e1482, 2023.

\bibitem{Xu2021LayoutLMv2MP}
Yang Xu, Yiheng Xu, Tengchao Lv, Lei Cui, Furu Wei, Guoxin Wang, Yijuan Lu, Dinei A.~F. Flor{\^e}ncio, Cha Zhang, Wanxiang Che, Min Zhang, and Lidong Zhou.
\newblock Layoutlmv2: Multi-modal pre-training for visually-rich document understanding.
\newblock In {\em ACL/IJCNLP}, 2021.

\bibitem{higashino1986knowledge}
Jun'ichi Higashino, Hiromichi Fujisawa, Yasuaki Nakano, and Masakazu Ejiri.
\newblock A knowledge-based segmentation method for document understanding.
\newblock {\em space}, 50(60):10, 1986.

\bibitem{wang1994analysis}
Dacheng Wang and Sargur~N Srihari.
\newblock Analysis of form images.
\newblock {\em International journal of pattern recognition and artificial intelligence}, 8(05):1031--1052, 1994.

\bibitem{tang1996automatic-document-processing-survey}
Yuan~Y. Tang, Seong-Whan Lee, and Ching~Y. Suen.
\newblock Automatic document processing: A survey.
\newblock {\em Pattern Recognition}, 29(12):1931--1952, 1996.

\bibitem{tang1993documentanalysisbook}
Yuan~Y. Tang, Chang~D. Yan, M.~Cheriet, and Ching~Y. Suen.
\newblock {\em Automatic Analysis and Understanding of Documents}, chapter 3.6, pages 625--654.
\newblock 1993.

\bibitem{Shoubin03023}
Shoubin Li, Xuyan Ma, Shuaiqun Pan, Jun Hu, Lin Shi, and Qing Wang.
\newblock Vtlayout: Fusion of visual and text features for document layout analysis.
\newblock In Duc~Nghia Pham, Thanaruk Theeramunkong, Guido Governatori, and Fenrong Liu, editors, {\em PRICAI 2021: Trends in Artificial Intelligence}, pages 308--322, Cham, 2021. Springer International Publishing.

\bibitem{zheng2023layoutdiffusion}
Guangcong Zheng, Xianpan Zhou, Xuewei Li, Zhongang Qi, Ying Shan, and Xi~Li.
\newblock Layoutdiffusion: Controllable diffusion model for layout-to-image generation, 2023.

\bibitem{oral2022fusion}
Berke Oral and G{\"u}l{\c{s}}en Eryi{\u{g}}it.
\newblock Fusion of visual representations for multimodal information extraction from unstructured transactional documents.
\newblock {\em International Journal on Document Analysis and Recognition (IJDAR)}, 25(3):187--205, 2022.

\bibitem{wu2016googles}
Yonghui Wu, Mike Schuster, Zhifeng Chen, Quoc~V. Le, Mohammad Norouzi, Wolfgang Macherey, Maxim Krikun, Yuan Cao, Qin Gao, Klaus Macherey, Jeff Klingner, Apurva Shah, Melvin Johnson, Xiaobing Liu, Łukasz Kaiser, Stephan Gouws, Yoshikiyo Kato, Taku Kudo, Hideto Kazawa, Keith Stevens, George Kurian, Nishant Patil, Wei Wang, Cliff Young, Jason Smith, Jason Riesa, Alex Rudnick, Oriol Vinyals, Greg Corrado, Macduff Hughes, and Jeffrey Dean.
\newblock Google's neural machine translation system: Bridging the gap between human and machine translation, 2016.

\bibitem{fu-etal-2019-graphrel}
Tsu-Jui Fu, Peng-Hsuan Li, and Wei-Yun Ma.
\newblock {G}raph{R}el: Modeling text as relational graphs for joint entity and relation extraction.
\newblock In {\em Proceedings of the 57th Annual Meeting of the Association for Computational Linguistics}, pages 1409--1418, Florence, Italy, July 2019. Association for Computational Linguistics.

\bibitem{pham2023deep}
Phu Pham, Loan~TT Nguyen, Witold Pedrycz, and Bay Vo.
\newblock Deep learning, graph-based text representation and classification: a survey, perspectives and challenges.
\newblock {\em Artificial Intelligence Review}, 56(6):4893--4927, 2023.

\bibitem{yu2020pick}
Wenwen Yu, Ning Lu, Xianbiao Qi, Ping Gong, and Rong Xiao.
\newblock Pick: Processing key information extraction from documents using improved graph learning-convolutional networks, 2020.

\bibitem{vaswani2017attention}
Ashish Vaswani, Noam Shazeer, Niki Parmar, Jakob Uszkoreit, Llion Jones, Aidan~N Gomez, {\L}ukasz Kaiser, and Illia Polosukhin.
\newblock Attention is all you need.
\newblock {\em Advances in neural information processing systems}, 30, 2017.

\bibitem{he2015deep}
Kaiming He, Xiangyu Zhang, Shaoqing Ren, and Jian Sun.
\newblock Deep residual learning for image recognition, 2015.

\bibitem{zhang2020trie}
Peng Zhang, Yunlu Xu, Zhanzhan Cheng, Shiliang Pu, Jing Lu, Liang Qiao, Yi~Niu, and Fei Wu.
\newblock Trie: End-to-end text reading and information extraction for document understanding, 2020.

\bibitem{xie2017aggregated}
Saining Xie, Ross Girshick, Piotr Dollár, Zhuowen Tu, and Kaiming He.
\newblock Aggregated residual transformations for deep neural networks, 2017.

\bibitem{lin2017feature}
Tsung-Yi Lin, Piotr Dollár, Ross Girshick, Kaiming He, Bharath Hariharan, and Serge Belongie.
\newblock Feature pyramid networks for object detection, 2017.

\bibitem{he2018mask}
Kaiming He, Georgia Gkioxari, Piotr Doll{\'a}r, and Ross Girshick.
\newblock Mask r-cnn.
\newblock In {\em Proceedings of the IEEE international conference on computer vision}, pages 2961--2969, 2017.

\bibitem{xu2023multimodal}
Peng Xu, Xiatian Zhu, and David~A. Clifton.
\newblock Multimodal learning with transformers: A survey, 2023.

\bibitem{feng2022multimodal}
Chun-Mei Feng, Yunlu Yan, Geng Chen, Yong Xu, Ling Shao, and Huazhu Fu.
\newblock Multi-modal transformer for accelerated mr imaging, 2022.

\bibitem{wang2023mutualformer}
Xixi Wang, Xiao Wang, Bo~Jiang, Jin Tang, and Bin Luo.
\newblock Mutualformer: Multi-modality representation learning via cross-diffusion attention, 2023.

\bibitem{li2021selfdoc}
Peizhao Li, Jiuxiang Gu, Jason Kuen, Vlad~I. Morariu, Handong Zhao, Rajiv Jain, Varun Manjunatha, and Hongfu Liu.
\newblock Selfdoc: Self-supervised document representation learning, 2021.

\bibitem{ren2015faster}
Shaoqing Ren, Kaiming He, Ross Girshick, and Jian Sun.
\newblock Faster r-cnn: Towards real-time object detection with region proposal networks.
\newblock {\em Advances in neural information processing systems}, 28, 2015.

\bibitem{mathew2021docvqa}
Minesh Mathew, Dimosthenis Karatzas, and CV~Jawahar.
\newblock Docvqa: A dataset for vqa on document images.
\newblock In {\em Proceedings of the IEEE/CVF Winter Conference on Applications of Computer Vision}, pages 2200--2209, 2021.

\bibitem{bao2020unilmv2}
Hangbo Bao, Li~Dong, Furu Wei, Wenhui Wang, Nan Yang, Xiaodong Liu, Yu~Wang, Songhao Piao, Jianfeng Gao, Ming Zhou, and Hsiao-Wuen Hon.
\newblock Unilmv2: Pseudo-masked language models for unified language model pre-training, 2020.

\bibitem{devlin2019bert}
Jacob Devlin, Ming-Wei Chang, Kenton Lee, and Kristina Toutanova.
\newblock Bert: Pre-training of deep bidirectional transformers for language understanding, 2019.

\bibitem{appalaraju2021docformer}
Srikar Appalaraju, Bhavan Jasani, Bhargava~Urala Kota, Yusheng Xie, and R.~Manmatha.
\newblock Docformer: End-to-end transformer for document understanding, 2021.

\bibitem{li2021structext}
Yulin Li, Yuxi Qian, Yuchen Yu, Xiameng Qin, Chengquan Zhang, Yan Liu, Kun Yao, Junyu Han, Jingtuo Liu, and Errui Ding.
\newblock Structext: Structured text understanding with multi-modal transformers, 2021.

\bibitem{hong2021bros}
Teakgyu Hong, Donghyun Kim, Mingi Ji, Wonseok Hwang, Daehyun Nam, and Sungrae Park.
\newblock Bros: A pre-trained language model focusing on text and layout for better key information extraction from documents, 2021.

\bibitem{gu2023unified}
Jiuxiang Gu, Ani~Nenkova Nenkova, Nikolaos Barmpalios, Vlad~Ion Morariu, Tong Sun, Rajiv~Bhawanji Jain, Jason Wen~Yong Kuen, and Handong Zhao.
\newblock Unified pretraining framework for document understanding, May~18 2023.
\newblock US Patent App. 17/528,061.

\bibitem{lu2019vilbert}
Jiasen Lu, Dhruv Batra, Devi Parikh, and Stefan Lee.
\newblock Vilbert: Pretraining task-agnostic visiolinguistic representations for vision-and-language tasks.
\newblock {\em Advances in neural information processing systems}, 32, 2019.

\bibitem{gu2020self}
Jiuxiang Gu, Jason Kuen, Shafiq Joty, Jianfei Cai, Vlad Morariu, Handong Zhao, and Tong Sun.
\newblock Self-supervised relationship probing.
\newblock {\em Advances in Neural Information Processing Systems}, 33:1841--1853, 2020.

\bibitem{chen2020simple}
Ting Chen, Simon Kornblith, Mohammad Norouzi, and Geoffrey Hinton.
\newblock A simple framework for contrastive learning of visual representations.
\newblock In {\em International conference on machine learning}, pages 1597--1607. PMLR, 2020.

\bibitem{powalski2021going}
Rafa{\l} Powalski, {\L}ukasz Borchmann, Dawid Jurkiewicz, Tomasz Dwojak, Micha{\l} Pietruszka, and Gabriela Pa{\l}ka.
\newblock Going full-tilt boogie on document understanding with text-image-layout transformer.
\newblock In {\em Document Analysis and Recognition--ICDAR 2021: 16th International Conference, Lausanne, Switzerland, September 5--10, 2021, Proceedings, Part II 16}, pages 732--747. Springer, 2021.

\bibitem{hewlett2016wikireading}
Daniel Hewlett, Alexandre Lacoste, Llion Jones, Illia Polosukhin, Andrew Fandrianto, Jay Han, Matthew Kelcey, and David Berthelot.
\newblock Wikireading: A novel large-scale language understanding task over wikipedia.
\newblock {\em arXiv preprint arXiv:1608.03542}, 2016.

\bibitem{raffel2020exploring}
Colin Raffel, Noam Shazeer, Adam Roberts, Katherine Lee, Sharan Narang, Michael Matena, Yanqi Zhou, Wei Li, and Peter~J Liu.
\newblock Exploring the limits of transfer learning with a unified text-to-text transformer.
\newblock {\em The Journal of Machine Learning Research}, 21(1):5485--5551, 2020.

\bibitem{ethayarajh2019contextual}
Kawin Ethayarajh.
\newblock How contextual are contextualized word representations? comparing the geometry of bert, elmo, and gpt-2 embeddings.
\newblock {\em arXiv preprint arXiv:1909.00512}, 2019.

\bibitem{dalmia2019enforcing}
Siddharth Dalmia, Abdelrahman Mohamed, Mike Lewis, Florian Metze, and Luke Zettlemoyer.
\newblock Enforcing encoder-decoder modularity in sequence-to-sequence models, 2019.

\bibitem{michael2019evaluating}
Johannes Michael, Roger Labahn, Tobias Grüning, and Jochen Zöllner.
\newblock Evaluating sequence-to-sequence models for handwritten text recognition, 2019.

\bibitem{farina2023distillation}
Marco Farina, Duccio Pappadopulo, Anant Gupta, Leslie Huang, Ozan İrsoy, and Thamar Solorio.
\newblock Distillation of encoder-decoder transformers for sequence labelling, 2023.

\bibitem{fu2023decoderonly}
Zihao Fu, Wai Lam, Qian Yu, Anthony Man-Cho So, Shengding Hu, Zhiyuan Liu, and Nigel Collier.
\newblock Decoder-only or encoder-decoder? interpreting language model as a regularized encoder-decoder, 2023.

\bibitem{feng2023sequence}
Shuwei Feng, Tianyang Zhan, Zhanming Jie, Trung~Quoc Luong, and Xiaoran Jin.
\newblock Sequence-to-sequence pre-training with unified modality masking for visual document understanding.
\newblock {\em arXiv preprint arXiv:2305.10448}, 2023.

\bibitem{peng2022ernie}
Qiming Peng, Yinxu Pan, Wenjin Wang, Bin Luo, Zhenyu Zhang, Zhengjie Huang, Teng Hu, Weichong Yin, Yongfeng Chen, Yin Zhang, et~al.
\newblock Ernie-layout: Layout knowledge enhanced pre-training for visually-rich document understanding.
\newblock {\em arXiv preprint arXiv:2210.06155}, 2022.

\bibitem{bao2022vlmo}
Hangbo Bao, Wenhui Wang, Li~Dong, Qiang Liu, Owais~Khan Mohammed, Kriti Aggarwal, Subhojit Som, Songhao Piao, and Furu Wei.
\newblock Vlmo: Unified vision-language pre-training with mixture-of-modality-experts.
\newblock {\em Advances in Neural Information Processing Systems}, 35:32897--32912, 2022.

\bibitem{appalaraju2023docformerv2}
Srikar Appalaraju, Peng Tang, Qi~Dong, Nishant Sankaran, Yichu Zhou, and R~Manmatha.
\newblock Docformerv2: Local features for document understanding.
\newblock {\em arXiv preprint arXiv:2306.01733}, 2023.

\bibitem{chen2020transformer}
Haifeng Chen, Dongmei Jiang, and Hichem Sahli.
\newblock Transformer encoder with multi-modal multi-head attention for continuous affect recognition.
\newblock {\em IEEE Transactions on Multimedia}, 23:4171--4183, 2020.

\bibitem{li2022two}
Bingjia Li, Jie Wang, Minyi Zhao, and Shuigeng Zhou.
\newblock Two-stage multimodality fusion for high-performance text-based visual question answering.
\newblock In {\em Proceedings of the Asian Conference on Computer Vision}, pages 4143--4159, 2022.

\bibitem{lee2022formnet}
Chen-Yu Lee, Chun-Liang Li, Timothy Dozat, Vincent Perot, Guolong Su, Nan Hua, Joshua Ainslie, Renshen Wang, Yasuhisa Fujii, and Tomas Pfister.
\newblock Formnet: Structural encoding beyond sequential modeling in form document information extraction, 2022.

\bibitem{hsiao2019flat2layout}
Chi-Wei Hsiao, Cheng Sun, Min Sun, and Hwann-Tzong Chen.
\newblock Flat2layout: Flat representation for estimating layout of general room types, 2019.

\bibitem{tang2023unifying}
Zineng Tang, Ziyi Yang, Guoxin Wang, Yuwei Fang, Yang Liu, Chenguang Zhu, Michael Zeng, Cha Zhang, and Mohit Bansal.
\newblock Unifying vision, text, and layout for universal document processing, 2023.

\bibitem{wang2022lilt}
Jiapeng Wang, Lianwen Jin, and Kai Ding.
\newblock Lilt: A simple yet effective language-independent layout transformer for structured document understanding, 2022.

\bibitem{voutharoja2023language}
Bhanu~Prakash Voutharoja, Lizhen Qu, and Fatemeh Shiri.
\newblock Language independent neuro-symbolic semantic parsing for form understanding, 2023.

\bibitem{gao2023enabling}
Tianyu Gao, Howard Yen, Jiatong Yu, and Danqi Chen.
\newblock Enabling large language models to generate text with citations, 2023.

\bibitem{nogueira2020navigation}
Rodrigo Nogueira, Zhiying Jiang, Kyunghyun Cho, and Jimmy Lin.
\newblock Navigation-based candidate expansion and pretrained language models for citation recommendation.
\newblock {\em Scientometrics}, 125:3001--3016, 2020.

\bibitem{Shi_2023_WACV}
Yuzhi Shi, Mijung Kim, and Yeongnam Chae.
\newblock Multi-scale cell-based layout representation for document understanding.
\newblock In {\em Proceedings of the IEEE/CVF Winter Conference on Applications of Computer Vision (WACV)}, pages 3670--3679, January 2023.

\bibitem{li-etal-2021-structurallm}
Chenliang Li, Bin Bi, Ming Yan, Wei Wang, Songfang Huang, Fei Huang, and Luo Si.
\newblock {S}tructural{LM}: Structural pre-training for form understanding.
\newblock In {\em Proceedings of the 59th Annual Meeting of the Association for Computational Linguistics and the 11th International Joint Conference on Natural Language Processing (Volume 1: Long Papers)}, pages 6309--6318, Online, August 2021. Association for Computational Linguistics.

\bibitem{yu2023structextv2}
Yuechen Yu, Yulin Li, Chengquan Zhang, Xiaoqiang Zhang, Zengyuan Guo, Xiameng Qin, Kun Yao, Junyu Han, Errui Ding, and Jingdong Wang.
\newblock Structextv2: Masked visual-textual prediction for document image pre-training.
\newblock {\em arXiv preprint arXiv:2303.00289}, 2023.

\bibitem{zhai2023fast}
Mingliang Zhai, Yulin Li, Xiameng Qin, Chen Yi, Qunyi Xie, Chengquan Zhang, Kun Yao, Yuwei Wu, and Yunde Jia.
\newblock Fast-structext: An efficient hourglass transformer with modality-guided dynamic token merge for document understanding.
\newblock {\em arXiv preprint arXiv:2305.11392}, 2023.

\bibitem{bao2021beit}
Hangbo Bao, Li~Dong, Songhao Piao, and Furu Wei.
\newblock Beit: Bert pre-training of image transformers.
\newblock {\em arXiv preprint arXiv:2106.08254}, 2021.

\bibitem{chen2022context}
Xiaokang Chen, Mingyu Ding, Xiaodi Wang, Ying Xin, Shentong Mo, Yunhao Wang, Shumin Han, Ping Luo, Gang Zeng, and Jingdong Wang.
\newblock Context autoencoder for self-supervised representation learning.
\newblock {\em arXiv preprint arXiv:2202.03026}, 2022.

\bibitem{wang2022rtformer}
Jian Wang, Chenhui Gou, Qiman Wu, Haocheng Feng, Junyu Han, Errui Ding, and Jingdong Wang.
\newblock Rtformer: Efficient design for real-time semantic segmentation with transformer.
\newblock {\em Advances in Neural Information Processing Systems}, 35:7423--7436, 2022.

\bibitem{harley2015icdar}
Adam~W Harley, Alex Ufkes, and Konstantinos~G Derpanis.
\newblock Evaluation of deep convolutional nets for document image classification and retrieval.
\newblock In {\em International Conference on Document Analysis and Recognition ({ICDAR})}.

\bibitem{cdip2021ftpdata}
David~D. Lewis, Gady Agam, Shlomo~Engelson Argamon, Ophir Frieder, David~A. Grossman, and Jefferson Heard.
\newblock Cdip dataset.
\newblock \url{https://ir.nist.gov/cdip/}, 2011.

\bibitem{schmidt2002building}
Heidi Schmidt, Karen Butter, and Cynthia Rider.
\newblock Building digital tobacco industry document libraries at the university of california, san francisco library/center for knowledge management.
\newblock {\em D-Lib Magazine}, 8(9):1082--9873, 2002.

\bibitem{lewis:iit-cdip-readme:2021}
David~D. Lewis, Gady Agam, Shlomo~Engelson Argamon, Ophir Frieder, David~A. Grossman, and Jefferson Heard.
\newblock The iit cdip test collection.
\newblock \url{https://ir.nist.gov/cdip/README.txt}, 2021.

\bibitem{jaume2019}
Jean-Philippe~Thiran Guillaume~Jaume, Hazim Kemal~Ekenel.
\newblock Funsd: A dataset for form understanding in noisy scanned documents.
\newblock In {\em Accepted to ICDAR-OST}, 2019.

\bibitem{vu2020revising}
Hieu~M. Vu and Diep Thi-Ngoc Nguyen.
\newblock Revising funsd dataset for key-value detection in document images, 2020.

\bibitem{xu2021layoutxlm}
Yiheng Xu, Tengchao Lv, Lei Cui, Guoxin Wang, Yijuan Lu, Dinei Florencio, Cha Zhang, and Furu Wei.
\newblock Layoutxlm: Multimodal pre-training for multilingual visually-rich document understanding, 2021.

\bibitem{microsoft2022readapi}
{Microsoft}.
\newblock What is optical character recognition?
\newblock \url{https://docs.microsoft.com/en-us/azure/cognitive-services/computer-vision/overview-ocr#read-api}, 2022.

\bibitem{davis2019NAFgithub}
Brian Davis, Bryan Morse, Scott Cohen, Brian Price, and Chris Tensmeyer.
\newblock National archives forms dataset.
\newblock \url{https://github.com/herobd/NAF_dataset}, 2019.

\bibitem{shinyama2019pdfminer}
Yusuke Shinyama et~al.
\newblock Pdfminer.
\newblock \url{https://github.com/euske/pdfminer}, 2019.

\bibitem{Huang_2019}
Zheng Huang, Kai Chen, Jianhua He, Xiang Bai, Dimosthenis Karatzas, Shijian Lu, and C.~V. Jawahar.
\newblock Icdar2019 competition on scanned receipt ocr and information extraction.
\newblock {\em 2019 International Conference on Document Analysis and Recognition (ICDAR)}, Sep 2019.

\bibitem{park2019cord}
Seunghyun Park, Seung Shin, Bado Lee, Junyeop Lee, Jaeheung Surh, Minjoon Seo, and Hwalsuk Lee.
\newblock Cord: A consolidated receipt dataset for post-ocr parsing.
\newblock In {\em Workshop on Document Intelligence at NeurIPS 2019}, 2019.

\bibitem{parker:github-cord-release:2021}
Fabrizio Primerano.
\newblock Availability of full dataset.
\newblock \url{https://github.com/clovaai/cord/issues/4}, 2021.

\bibitem{mathew2021docvqadataset}
Minesh Mathew, Dimosthenis Karatzas, and CV~Jawahar.
\newblock Docvqa: A dataset for vqa on document images.
\newblock \url{https://www.docvqa.org/datasets}, 2021.

\bibitem{ucsflibrary:IndustryDocumentsLibrary}
{UCSF Library}.
\newblock Industry documents library.
\newblock \url{https://www.industrydocuments.ucsf.edu/}.

\bibitem{ding2023form}
Yihao Ding, Siqu Long, Jiabin Huang, Kaixuan Ren, Xingxiang Luo, Hyunsuk Chung, and Soyeon~Caren Han.
\newblock Form-nlu: Dataset for the form language understanding.
\newblock {\em arXiv preprint arXiv:2304.01577}, 2023.

\bibitem{biten2019scene}
Ali~Furkan Biten, Ruben Tito, Andres Mafla, Lluis Gomez, Marçal Rusiñol, Ernest Valveny, C.~V. Jawahar, and Dimosthenis Karatzas.
\newblock Scene text visual question answering, 2019.

\bibitem{li2022dit}
Junlong Li, Yiheng Xu, Tengchao Lv, Lei Cui, Cha Zhang, and Furu Wei.
\newblock Dit: Self-supervised pre-training for document image transformer.
\newblock In {\em Proceedings of the 30th ACM International Conference on Multimedia}, pages 3530--3539, 2022.

\end{thebibliography}
\bibliographystyle{unsrt}


\end{document}